%% file: Sorometry.tex
\documentclass[a4paper,12pt]{article}
\usepackage[pdftex]{graphicx} % Required for inserting images

\usepackage{graphicx}%
\usepackage{geometry}
\usepackage[utf8]{inputenc}
\usepackage{makecell}
\usepackage{multirow}%
\usepackage{amsmath,amssymb,amsfonts}%
\usepackage{amsthm}%
\usepackage{rotating}
\usepackage{mathrsfs}%
\usepackage[title]{appendix}%
\usepackage{hyperref}
\usepackage{url}
\usepackage{svg}
\usepackage{natbib}
\usepackage{textcomp}%
\usepackage{manyfoot}%
\usepackage{booktabs}%
\usepackage{algorithm}%
\usepackage{algorithmicx}%
\usepackage{algpseudocode}%
\usepackage{listings}%
\usepackage{enumitem}
\usepackage{adjustbox}
\usepackage{rotating}
\usepackage[affil-it]{authblk}
\usepackage{lipsum}
\usepackage{longtable}
\usepackage{lscape}
\usepackage{tikz}
\usepackage{xcolor}%
\usepackage{graphicx}
\usetikzlibrary{arrows.meta,positioning}

\pgfdeclarelayer{bg}
\pgfsetlayers{bg,main}
\definecolor{stepgreen}{RGB}{51,255,102}
\definecolor{datablue}{RGB}{51,221,255}

\hypersetup{
    colorlinks = true,
    allcolors = blue
}
\usepackage[many]{tcolorbox}

\begin{document}
 \pagenumbering{gobble}
 
\title{\vspace*{-0.5in} \textbf{\Large Leveraging Phytolith Research using \\Artificial Intelligence}}
\author[1*]{\normalsize Andr\'es G. Mej\'ia Ram\'on}
%\email{Andres.Mejia@OIST.jp}
\author[1]{\normalsize Kate Dudgeon}
\author[1]{\normalsize Nina Witteveen}
\author[2,3]{\normalsize Dolores Piperno}
\author[4]{\\ \normalsize Michael Kloster}
\author[5]{\normalsize Luigi Palopoli}
\author[6]{\normalsize M\'onica Moraes R.}
\author[7]{\normalsize Jos\'e M. Capriles}
\author[1,8]{\\ \normalsize Umberto Lombardo}
%\email{munson@lycoming.edu}
\affil[1]{\footnotesize \textit{Institut de Ci\`encia i Tecnologia Ambientals, Universitat Aut\`onoma de Barcelona} \authorcr \textit{Carrer de les Columnes s/n, Cerdanyola del Vall\`es, Barcelona, 08193, Espanya} \vspace*{6pt}}
\affil[2]{\footnotesize \textit{Department of Anthropology, Smithsonian National Museum of Natural History,} \authorcr \textit{1000 Madison Drive NW, Washington, D.C., 20560, United States}\vspace*{6pt}}
\affil[3]{\footnotesize \textit{Smithsonian Tropical Research Institute,} \authorcr \textit{Luis Clement Avenue, Bldg. 401 Tupper, Anc\'on, Ciudad de Panam\'a, Rep\'ublica de Panam\'a}
\vspace*{6pt}}
\affil[4]{\footnotesize \textit{Phycology Group, Faculty of Biology, University of Duisburg-Essen,} \authorcr \textit{Universitätsstr. 2, 45141 Essen, Deutschland}
\vspace*{6pt}}
\affil[5]{\footnotesize \textit{Dipartimento di Ingegneria e Scienza dell’Informazione, Universit\`a di Trento} \authorcr \textit{Via Sommarive 9, 38123, Trento, Italia}
\vspace*{6pt}}
\affil[6]{\footnotesize \textit{Herbario Nacional de Bolivia, Instituto de Ecolog\'ia, Universidad Mayor de San Andr\'es,} \authorcr \textit{Calle 27 y Andrés Bello s/n, Cota Cota, La Paz, Bolivia} \vspace*{6pt}}
\affil[7]{\footnotesize \textit{Department of Anthropology, The Pennsylvania State University,} \authorcr \textit{Susan Welch Liberal Arts Building, University Park, Pennsylvania, 16803, United States} \vspace*{6pt}}
\affil[8]{\footnotesize \textit{Departament de Prehist\`oria, Universitat Aut\`onoma de Barcelona,} \authorcr \textit{Edifici B, Carrer de la Fortuna, Cerdanyola del Vall\`es, Barcelona, 08193, Espanya} \vspace*{6pt}}
\affil[*]{\footnotesize Corresponding author: Andres.Mejia@UAB.cat}

\maketitle

\date{\vspace{-5ex}}

\begin{abstract}
    \noindent \footnotesize Phytolith analysis is a crucial tool for reconstructing past vegetation and human activities, but traditional methods are severely limited by labour-intensive, time-consuming manual microscopy. To address this bottleneck, we present Sorometry: a comprehensive end-to-end artificial intelligence pipeline for the high-throughput digitisation, inference, and interpretation of phytoliths. Our workflow processes z-stacked optical microscope scans to automatically generate synchronised 2D orthoimages and 3D point clouds of individual microscopic particles. We developed a multimodal fusion model that combines ConvNeXt for 2D image analysis and PointNet++ for 3D point cloud analysis, supported by a graphical user interface for expert annotation and review. Tested on reference collections and archaeological samples from the Bolivian Amazon, our fusion model achieved a global classification accuracy of 77.9\% across 24 diagnostic morphotypes and 84.5\% for segmentation quality. Crucially, the integration of 3D data proved essential for distinguishing complex morphotypes (such as grass silica short cell phytoliths) whose diagnostic features are often obscured by their orientation in 2D projections. Beyond individual object classification, Sorometry incorporates Bayesian finite mixture modelling to predict overall plant source contributions at the assemblage level, successfully identifying specific plants like maize and palms in complex mixed samples. This integrated platform transforms phytolith research into an ``omics"-scale discipline, dramatically expanding analytical capacity, standardising expert judgements, and enabling reproducible, population-level characterisations of archaeological and paleoecological assemblages. \\

\end{abstract}

\normalsize 

\section{Introduction}
Phytoliths are durable microscopic silicate particles produced by plants that persist in a diverse range of soils and sediments over long time scales (\citealt{piperno_phytoliths_2006}). Phytoliths can be used to reconstruct past vegetation change and past human activities at millenary timescales (\citealt{ball_phytoliths_2016,chen_review_2024,hodson_development_2016,neumann_phytoliths_2017,stromberg_functions_2016,witteveen_quantifying_2024}). Traditional phytolith analysis is labour intensive and time consuming. It requires a significant number of person-hours of observation and slide manipulation over a microscope---and commonly focuses on a small subset of all phytoliths present within a slide potentially neglecting significant useful information. The identification of a standard 200-400 diagnostic morphotypes in one sample can take several hours, limiting the total number of samples which can be feasibly analysed to address specific research questions. Although several crops produce diagnostic phytoliths (\citealt{piperno_phytoliths_2006,iriarte_assessing_2003}), most produce them in low amounts, making their presence in the paleoenvironmental record rare.

Phytoliths are commonly identified by their size, shape, and surface pattern. Phytolith research's reliance on pattern recognition in two- and three-dimensional representations makes it particularly well-suited for efficiency improvements using by leveraging modern advances in artificial intelligence (AI) and machine learning (ML)---See Supplementary Materials A for a non-technical glossary of key terms. Previous applications of AI in phytolith research focused primarily on distinguishing between a very limited set of morphotypes---for example eight morphotypes from 429 images (\citealt{Diez-Pastor2020}), or distinguishing one specific morphotype between closely-related taxa (\citealt{cai_machine_2017})---in highly controlled settings with limited applicability to complex soil phytolith assemblages, and required significant human intervention in scanning, processing, and labelling. These studies also relied exclusively on two-dimensional imagery, when many diagnostic features of particular morphotypes are three-dimensional in nature. Three-dimensional modelling to date has been limited to single-phytolith reconstructions using  time-intensive confocal microscopy, employing descriptive morphometrics manually obtained from the models in standard convolutional neural networks (\citealt{Gallaher2020}), greatly hampering its ability to generate datasets of a sufficient scale.

\begin{figure}
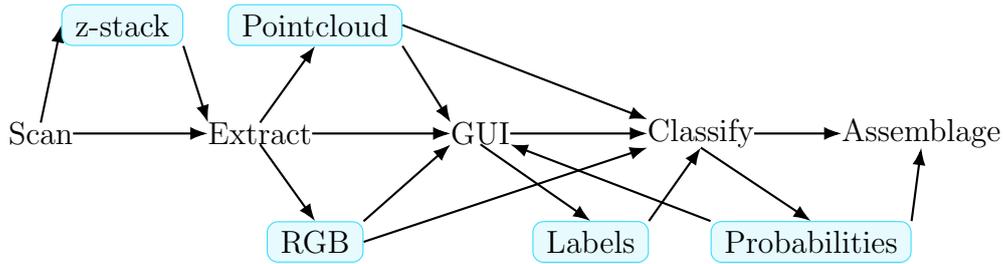

    \centering
    \include{Pipeline_Fig}
    \caption{Sorometry workflow pipeline, with associated data productsin blue.}
    \label{Pipeline_Fig}
\end{figure}

In this paper, we present Sorometry---a set of resources that leverage recent advances in artificial intelligence to permit phytolith analysis at scales orders of magnitude beyond traditional methods. We show the resources' utility (and that of AI in phytolith research more broadly) though a case study on samples extracted from archaeological sediments in the Bolivian Amazon. Sorometry covers all aspects of complete end-to-end pipeline (Figure \ref{Pipeline_Fig}) for phytolith slide digitisation, inference, and interpretation: (1) a data pre-processing pipeline to transform focus-stack imagery from digital slide scanning optical microscopes into segmented 3D point cloud and 2D image representations of the constituent phytoliths; (2) a suite of convolutional neural networks adapted to the output 2D and 3D data structures, with additional pre-trained weights based on (3) pairs of point clouds and images for 12,518 phytoliths with morphological labels for 4,114 of those made by phytolith experts, (4) images and pointclouds of 712 complete 2 mm x 2 mm sections of slides, (5) a graphical user interface (GUI) that allows users to label segmented objects, run analytical operations, and view model predictions, (6) additional analytical tools to permit assemblage-scale exploratory data analysis and mixture modelling. The use of modern AI technology--namely, the large language model (LLM) Codex---enabled us to raise the level of ambition of our tool-suite, transforming it from a simple in-house prototype to a user-friendly, robust and professional piece of software engineering.

\section{Results}

\subsection{Overview of the Sorometry workflow}
\begin{figure}[t!]
    \centering
    \includegraphics[width=\textwidth]{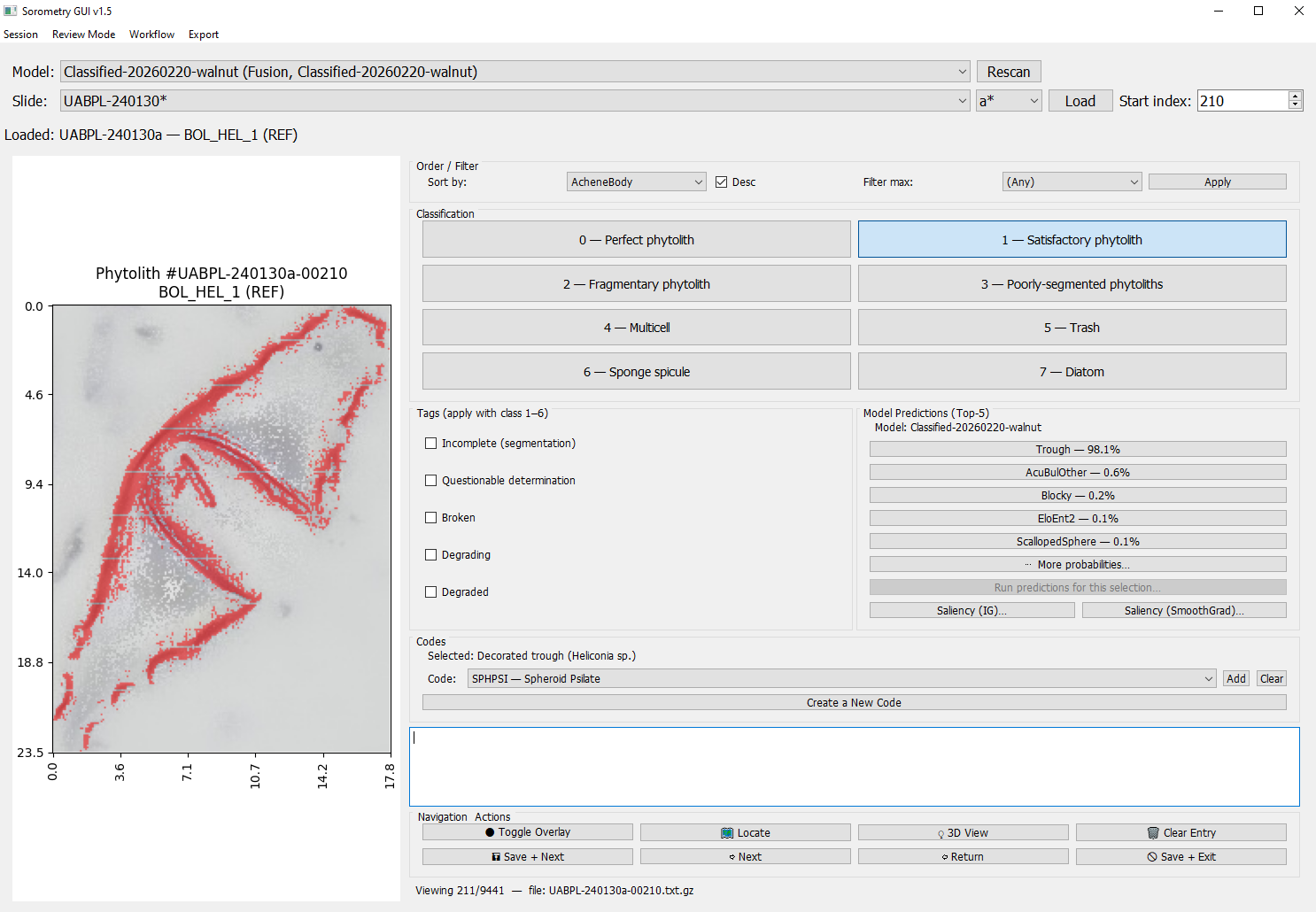}
    \caption{Sorometry Graphical User Interface.}
    \label{GUI}
\end{figure}

\begin{sidewaysfigure}
    \centering
    \includegraphics[width=1\paperwidth]{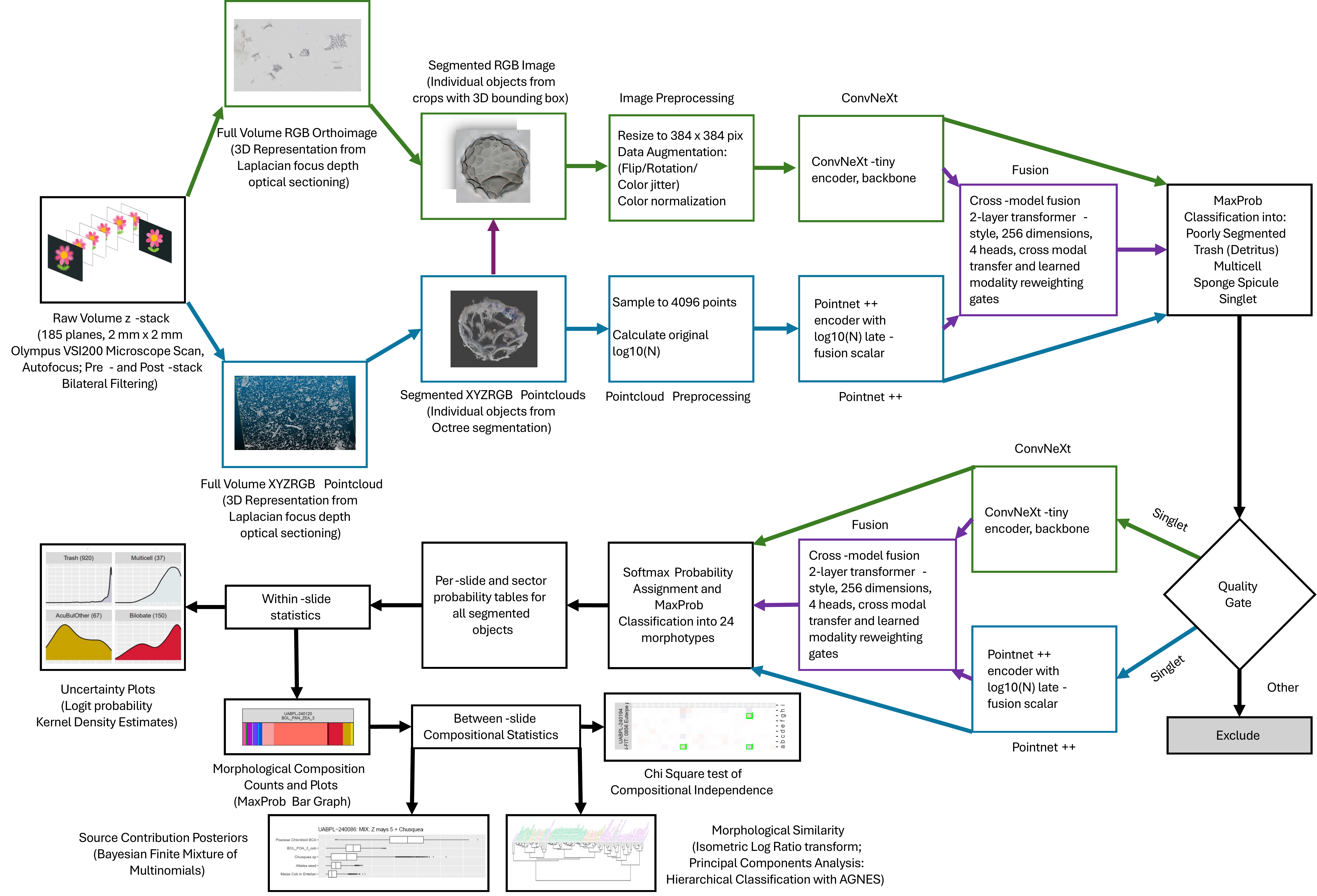}
    \caption{Data pipeline architecture, with the 2D image stream in green, the 3D pointcloud stream in blue, and fusion modality in purple.}
    \label{Architecture}
\end{sidewaysfigure}

Sorometry (from \textit{soro-} meaning `stack', and \textit{-metry} meaning `to measure') contains a sample-to-prediction-and-modelling pipeline (Figures \ref{Pipeline_Fig} - \ref{Architecture}), wrapped within a graphical user interface for large-scale data labelling and data processing (Figure \ref{GUI}). The first step is to scan samples from mounted slides using a protocol that maximises the number of non-overlapping phytoliths set on a single plane and the contrast between phytoliths and mounting medium to aid automatic detection \citealt{dudgeon_scaling_2026}). Slides are scanned using a digital microscope (Olympus VS200) at sufficient focal planes (‘z-slices') to image the lowermost 49 \textmu{}m of the slide where the majority of phytoliths have settled. Digital microscope are already routinely used in biomedical imaging work flows (Williams et al., 2023) and present the most cost-effective, rapid means of generating large, high quality training sets required for implementing automated workflows incorporating AI to locate and classify phytoliths at scale, and in experimental applications.

\begin{figure}[t!]
    \centering
    \includegraphics[width=0.3\linewidth]{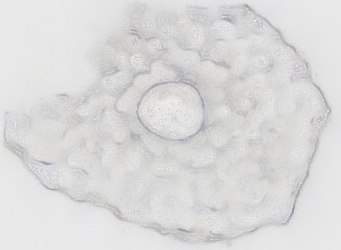}
    \hspace*{16pt}
    \includegraphics[width=0.3\linewidth]{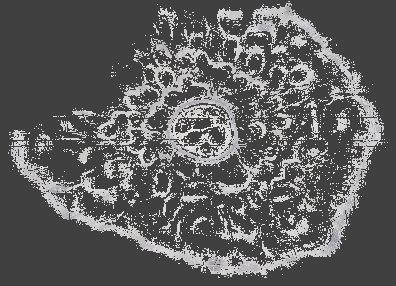}
    \caption{Two representations of the same Cyperaceae \textsc{Polygonal Achene Body} phytolith: (Left) 2-Dimensional RGB Orthoimage crop; (Right) 3-Dimensional Pointcloud}
    \label{Phytolith}
\end{figure}

After scanning, the two- and three-dimensional representations of the entire scanned volume are generated. A bilateral filter is first applied to each to each z-slice of the ‘z-stacks', reducing high-frequency noise while increasing low-frequency contrast particularly at the edges of objects. A Laplacian is then used to identify edges on each z-slice. Focus-stacked orthoimages are produced by selecting the colour value of the pixel with the highest Laplacian value. Since a point on an object will be in focus at the focal distance corresponding to its z-coordinate, and the x-y coordinates are given by a pixel's row and column, point clouds can be generated by retaining the coordinates of all pixels with a Laplacian value greater than background. Point clouds and images are segmented using octree-based segmentation on the point clouds alone, with the orthoimages cropped using the bounding boxes of the segmented point clouds (Figure \ref{Phytolith}).

Using the GUI, phytolith representations are presented to phytolith experts who can assign labels based on morphology, condition, and segmentation quality (Supplementary Material B). The experts are presented with a 2D image accompanied by measurements and overlain by an xy projection of the point cloud. The experts may toggle the overlay, rotate the raw point cloud in 3D, and find the original position of the phytolith on the full-scan image to assist their identification. If any predictions from a previously-trained CNN are available for that phytolith, the GUI suggests the highest-probability classes.

Once a labelled data set has been generated, sorometry can use them to train Pointnet-family CNNs (\citealt{Qi2017,qian_pointnext_2022}) with an added late-fusion scalar on the point clouds, a ConvNeXT transformer-style CNN (\citealt{liu_convnet_2022}) on the images, and a Fusion model of one Pointnet-family model and ConvNeXT (Figure \ref{Architecture}). Since segments can often contain extraction detritus and poorly segmented phytoliths, the pipeline first trains one model on segmentation quality and condition (i.e. detritus vs. phytolith, and poor segmentation vs. proper segmentation), using it as a pre-filter before predicting morphology. Both models can then be applied to the rest of the unlabeled dataset.

After an inference, sorometry enables the user to view summary statistics for the distribution of morphological predictions within slides to inform further slide analysis and object labelling. Bar charts indicate the percent composition of a scanned volume by maximum-probability morphotype, while kernel density plots for each predicted morphotype within a sector show the distribution of confidence values for all such morphotypes within the slide. If multiple volumes are scanned within the same slide, a two-tailed chi-square test of independence allows users to evaluate if their scanned samples are likely representative of the slide---with factor loadings indicating abundance or scarcity of particular predicted morphotypes---whose failure would indicate either biases in morphotype distribution on a slide or a problem in model classification (Figure 4).

Sorometry likewise provides a set of exploratory data analytical tools for comparison of morphological compositional data between slides, under the assumption that---even if certain predictions are incorrect---so long as they are systematically incorrect, empirical signals of morphological compositional similarity can still be extracted. These allow users to rapidly structure large datasets and identify slides to prioritise more in-depth, human analysis. We are employing this to identify slides that may contain culturally relevant plants even if we have yet to train a model on a relevant diagnostic morphotype. Barcode plots display colour-coded compositional bar graphs together, allowing users to visually identify potential similarities and outliers. Principal components analysis on an isometric-log ratio transform of the compositional data allows the user to plot slides and sectors in compositional space, while hierarchical clustering and dendrograms can be used identify samples with similar distributions of predicted morphotypes.

Finally, sorometry allows users to predict what plants may have contributed to a given assemblage through a Bayesian finite mixture of multinomials. This considers every plant to be a phytolith-generating ‘process', producing known relative distributions of predicted morphotypes. An archaeological sample would then be a collection of categorical counts (the number of each morphotype), resulting from a weighted mixture of phytolith-generating processes. The resulting probability distributions indicate the relative likelihood that a particular type of plant contributed to the mixture even if absent ‘diagnostic' morphotypes.

\subsection{Digitisation, Segmentation, and Labelling}
For the present study, we scanned a total of 712 2 mm x 2 mm areas  (‘sectors') of 123 slides, for an average of 6.75 sectors per slide. From these, three contexts are extracts of sediments from an archaeological excavation, nine are extracts of sediments from a soil core atop the archaeological excavation, and fifteen are extracts of known biological plant material from our reference collection (Supplementary Materials B). Segmentation of these sectors resulted in 3.81 million segmented point clouds, for an average of 5,219 segments per sector.

\begin{figure}
\centering
\includegraphics[width=\linewidth]{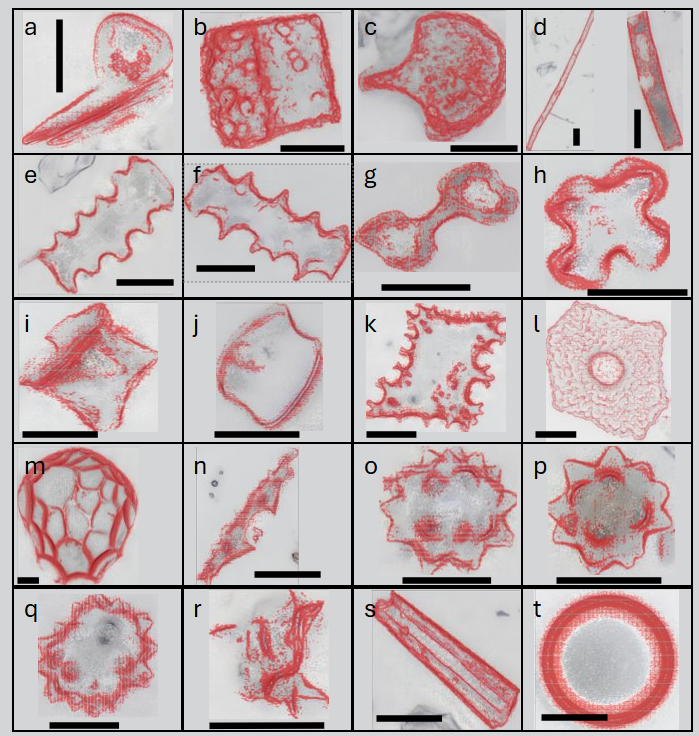}
\caption{Digital images of micro-fossils included in the study. (a) \textsc{Acute bulbosus}, ACUBUL; (b) \textsc{Blocky}, BLO; (c) \textsc{Bulliform flabellate}, BULFLA; (d) \textsc{Elongate entire}, ELOENT; (e) \textsc{Elongate sinuate}, ELOSIN; (f) Elongate dendritic/dentate, ELODET/DEN; (g) \textsc{Bilobate}, BIL; (h) \textsc{Cross}, CRO; (i) \textsc{Rondel}, RON; (j) \textsc{Saddle}, SAD; (k) Decorated tabular, TABDEC; (l) \textsc{Polygonal achene body}, POLACH; (m) Scalloped sphere, SSP; (n) \textsc{Trough}, TRO; (o) \textsc{Spheroid echinate}, SPHECH; (p) Spheroid symmetrical, SPHSYM; (q) \textsc{Ellipsoidal echinate}, ELLECH; (r) \textsc{Druse}, DRU; (s) Non-phytolith \textsc{Sponge spicule}, SPO; (t) Non-phytolith \textsc{Silica Microsphere}, \textsc{SilicaBall}. Scale bar: 10 \textmu{}m. The red overlay represents the automatically segmented portion of the phytolith based on the point-cloud extraction. Detailed descriptions of morphological criteria, sub-categories of diverse groups and rationale for inclusion in the study are provided in Supplementary Materials B.}
\label{Type_Images}
\end{figure}

Thus far, 15,842 phytoliths have been assigned a condition and segmentation quality ‘classification' , and of the well-segmented phytoliths 4638 have been labelled with a morphological type ‘code' (Tables \ref{AllClasses_Count} - \ref{Validation_Count}).  Of these, we selected 24 morphotypes with at least 20 labelled observations to include in the classification models (Figure \ref{Type_Images}), based on a number of practical considerations (Supplementary Materials B).

\subsection{Automated Classification}
To evaluate a model's performance, we can rely on the global accuracy (the percent of test observations correctly predicted), the class-ajusted accuracy (an arithmetic mean of the accuraccy of each class), and the Macro F1 score (a harmonic mean of the accuracy, penalising large deviations more than the class-adjusted accuracy). When predicting morphotypes, of the three tested single-data models ConvNeXT performed best with a global accuracy of 74.0\%, a class-adjusted accuracy of 65.6\%, and a Macro F1 score of 0.67. Of the two 3D models, the older Pointnet++ outperformed PointNeXt, and was thus used alongisde ConvNeXt in the fusion model. The Fusion model was the best overall model with a global accuracy of 77.9\%, a class-adjusted accuracy of 71.4\%, and a Macro F1 score of 0.71. Likewise, the fusion model was best in predicting segmentation quality, with a global accuracy of 84.5\%, a class-adjusted accuracy of 62.9\% and macro F1 score of 68.7\% and only an 8\% false negative rate for Well-Segmented objects. (Figure \ref{AllClasses_Fusion}).

\begin{figure}
\centering
\includegraphics[width=\linewidth]{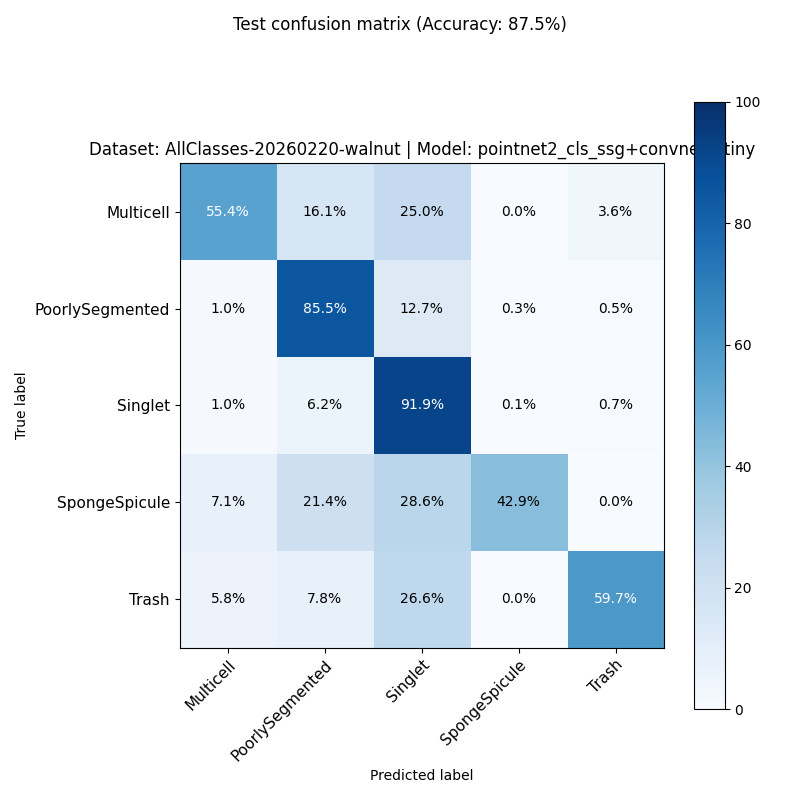}
\caption{Morphotype classification performance for the Fusion model, showing percent true values classified as a given predicted value.}
\label{AllClasses_Fusion}
\end{figure}

In contrast to the segmentation quality classification, per-morphotype accuracies depend more on the three-dimensional physiognomy of the phytolith. Accuracies and can be evaluated using confusion matrix. Rows represent an object's true classification, and columns the model's prediction, with the percent value indicating the percent of true values predicted to be the column's morphotype. Identification of the diagnostic morphotypes and non-phytolith objects was successful in all three models. All three Elongate categories performed better in the point cloud-based training sets compared with images. The point clouds  captured 3D morphology---occluded in 2D images---which could provide higher taxonomic resolution of Elongate forms in some species (Albert et al., 2003). \textsc{Elongate dentate/dendritic} (\textsc{EloDetDen}), \textsc{Elongate sinuate} (\textsc{EloSin}), and \textsc{ZMBac} both performed better in the ConvNeXT compared with Pointnet++ and Fusion models, suggesting the image-based classification is more sensitive to subtle variations in margin decoration compared with the point clouds, where the 3D morphology (e.g. thick or thin) is not important for classification.

\begin{figure}
\centering
\includegraphics[width=\linewidth]{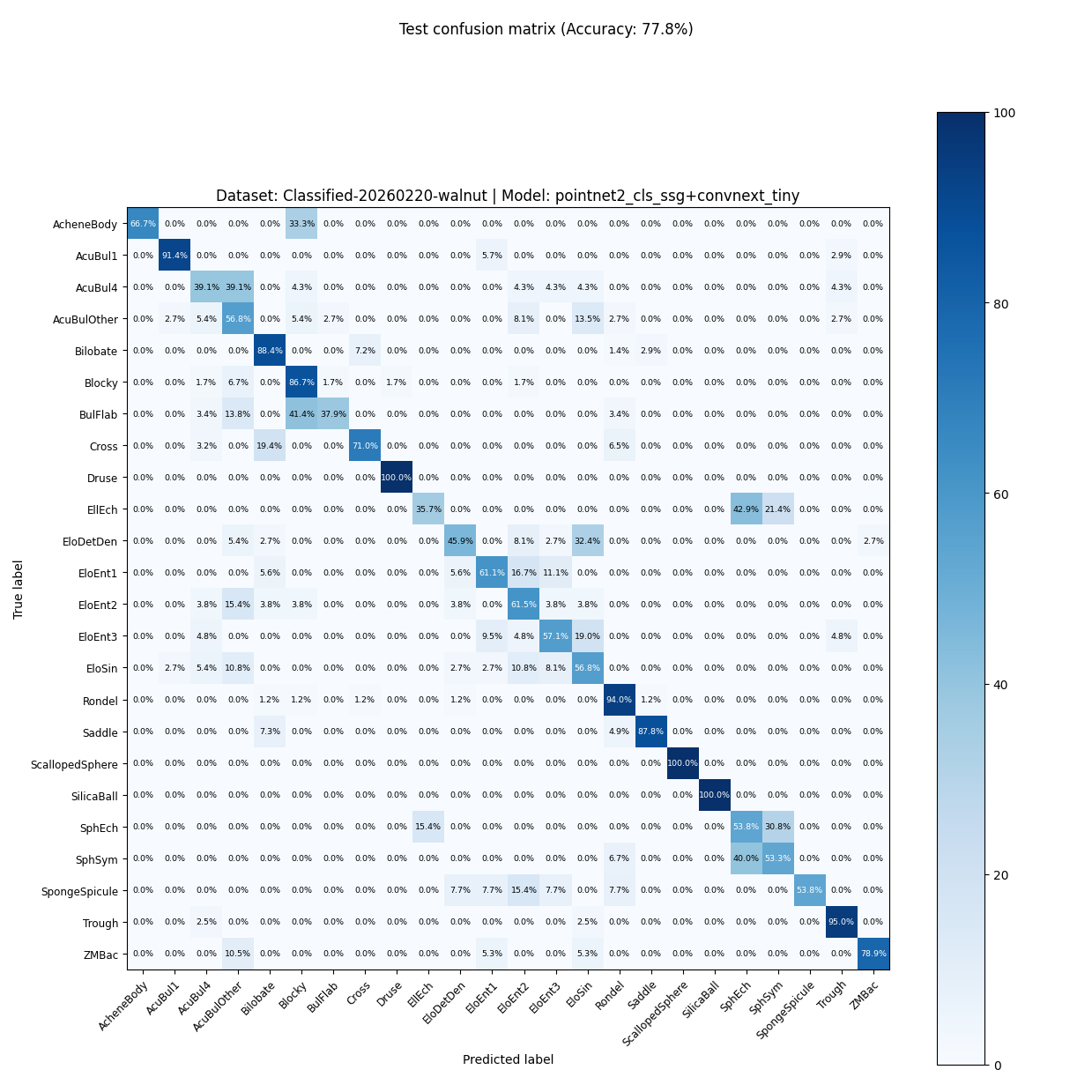}
\caption{Morphological classification performance for the Fusion model, showing percent true values classified as a given predicted value.}
\label{AllClasses_Fusion}
\end{figure}

Barring \textsc{Saddle}s, grass silica short cell phytoliths (GSSCPs) were all best classified by fusion model. The point clouds were often essential for distinguishing between GSSCPs, where diagnostic attributes were not visible because of the orientation of the phytolith in the slide, whilst the images contained different diagnostic information not always captured in the point clouds. The identification of palms phytoliths; \textsc{EllEch}, \textsc{SphEch} and, \textsc{SphSym}, showed high potential. As almost all the misidentifications occurred between these morphotypes, demonstrating a practical application where palms can be distinguished from other vegetation types in a soil phytolith assemblage. Moreover, \textsc{SphEch} and \textsc{SphSym}'s above-50\% accuracies, combined with \textsc{EllEch}'s lack of false positives make the model useful in surfacing objects that are highly likely to be such morphotypes and provide additional useful `spectral' resolution for an assemblage analysis so long as errors are consistent and systematic between similar objects. 

\begin{figure}
\centering
\includegraphics[width=\linewidth]{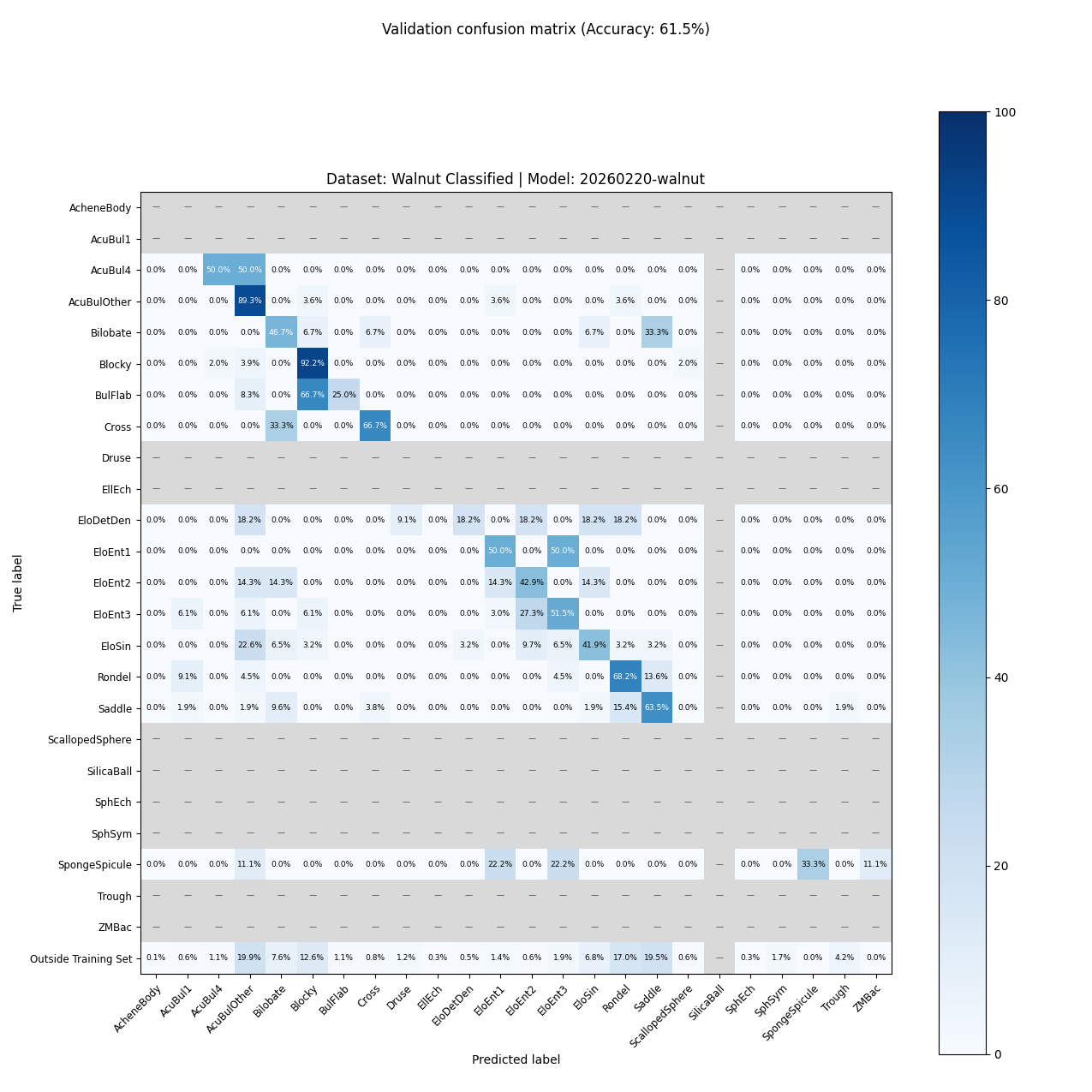}
\caption{Morphological classification performance for the Fusion model on well-segmented phytoliths against an archaeological validation slide, showing percent true values classified as a given predicted value.}
\label{Confusion_Archy}
\end{figure}

We also assessed the performance of the model against an archaeological slide with morphotypes not included in the training dataset (Figure \ref{Confusion_Archy}). Performance was predictably much poorer, but the results suggests a more differentiated picture than a simple drop in performance on an out-of-sample assemblage. Within the \textsc{Acute bulbosus} complex, all archaeological examples were recognised, and the residual confusion between \textsc{AcuBul4} and \textsc{AcuBulOther} is readily explained by the structure of the categories themselves: \textsc{AcuBul4} is comparatively discrete, whereas \textsc{AcuBulOther} is intentionally broad and includes transitional forms that approach \textsc{AcuBul4} morphologically, a problem compounded by still-limited counts for the sub-categories. 

The Elongate classes show a similar issue, but for different reasons. Misclassification among \textsc{EloEnt2} and \textsc{EloEnt3} is of limited interpretive consequence because their present separation rests on visible form rather than a firm taxonomic distinction, while confusion across \textsc{EloDetDen}, \textsc{EloSin}, and EloEnt more generally reflects the fact that these labels are currently assigned primarily by margin decoration even though other attributes—surface texture, proportions, thickness, and overall 3D form—can make specimens with different margins more similar than specimens sharing the same nominal class. This is especially true for degraded and transitional archaeological forms, including cases with ambiguous or mixed margins, where the model often elevated the two most plausible categories rather than failing arbitrarily. 

Performance among GSSCPs was likewise weaker than desirable, but part of this appears to stem from validation specimens that were almost certainly short-cell phytoliths even when image or point-cloud quality prevented secure assignment to \textsc{Bilobate}, \textsc{Rondel}, \textsc{Cross}, or \textsc{Saddle}, suggesting that the confusion matrix may understate practical sensitivity. Palms remain promising but undertrained, as the categories are morphologically close and still represented by only modest reference counts. Finally, objects outside the 24 trained morphotypes were generally absorbed into the largest and most internally variable classes rather than into the tightest, most distinctive categories, which is encouraging insofar as rare unknowns were seldom forced into sharply delimited morphotypes such as \textsc{AcuBul1}, achene body, scalloped sphere, or \textsc{ZMBac}. 

Taken together, these patterns indicate that the archaeological confusion matrix is capturing real morphological continua, uneven training coverage, and the presence of genuinely unmodeled forms, and that the clearest path forward is not merely more data, but larger and better-structured training sets, explicit treatment of degradation and transitional states, finer sub-categorisation of heterogeneous classes, and eventually a more quantitative morphometric basis for defining class boundaries and class-specific rejection thresholds.

\subsection{Assemblage Analysis}
\begin{figure}
    \centering
    \includegraphics[width=\linewidth]{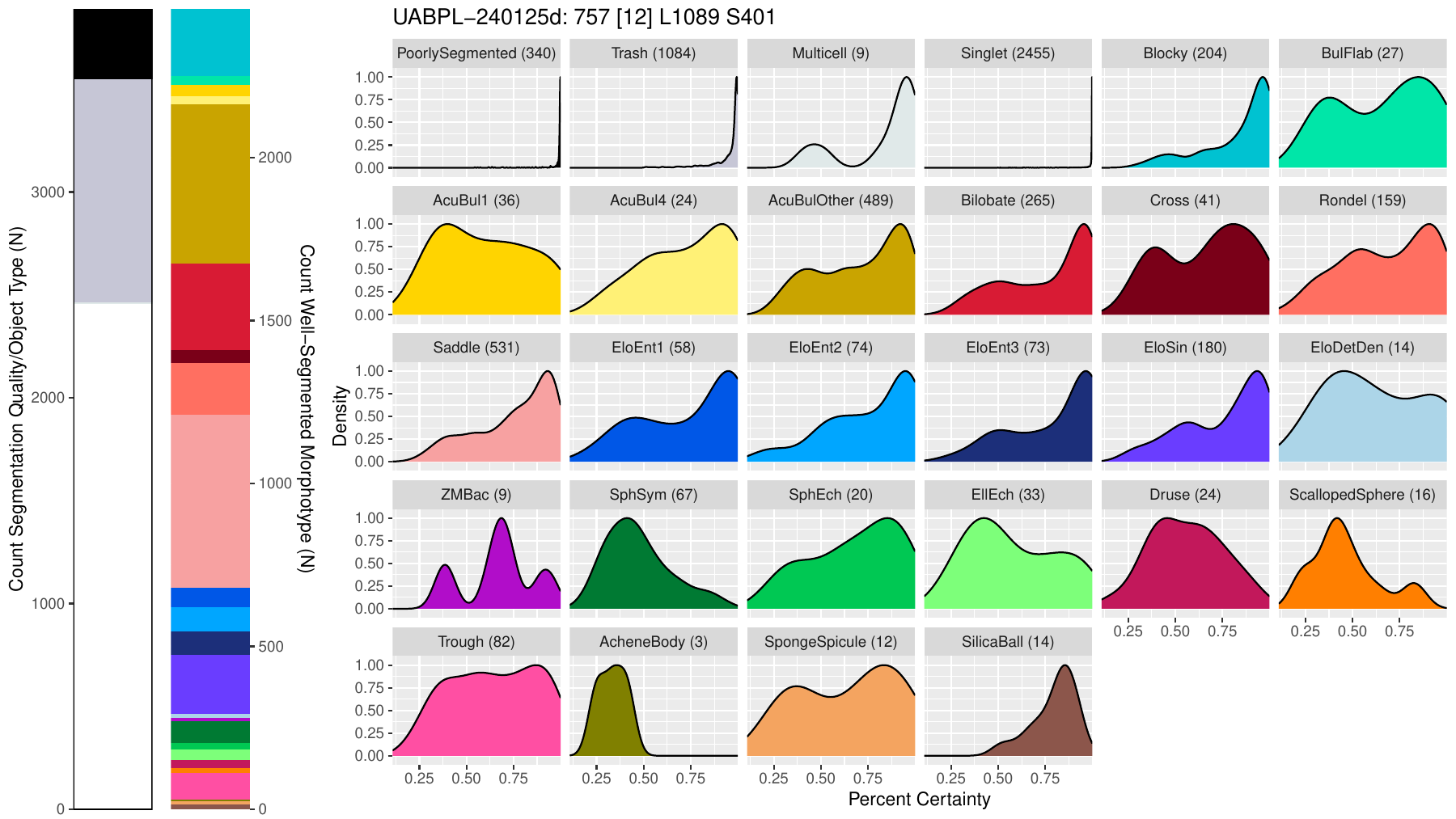}
    \caption{Composition diagram for volume UABPL-240125d. (Left) Bar graphs displaying the predicted number of objects for each category in each model. (Right) Kernel density estimate of the percent certainty for all objects given their maximum probability class.}
    \label{Uncertainty}
\end{figure}

\begin{sidewaysfigure}
    \centering
    \includegraphics[width=\linewidth]{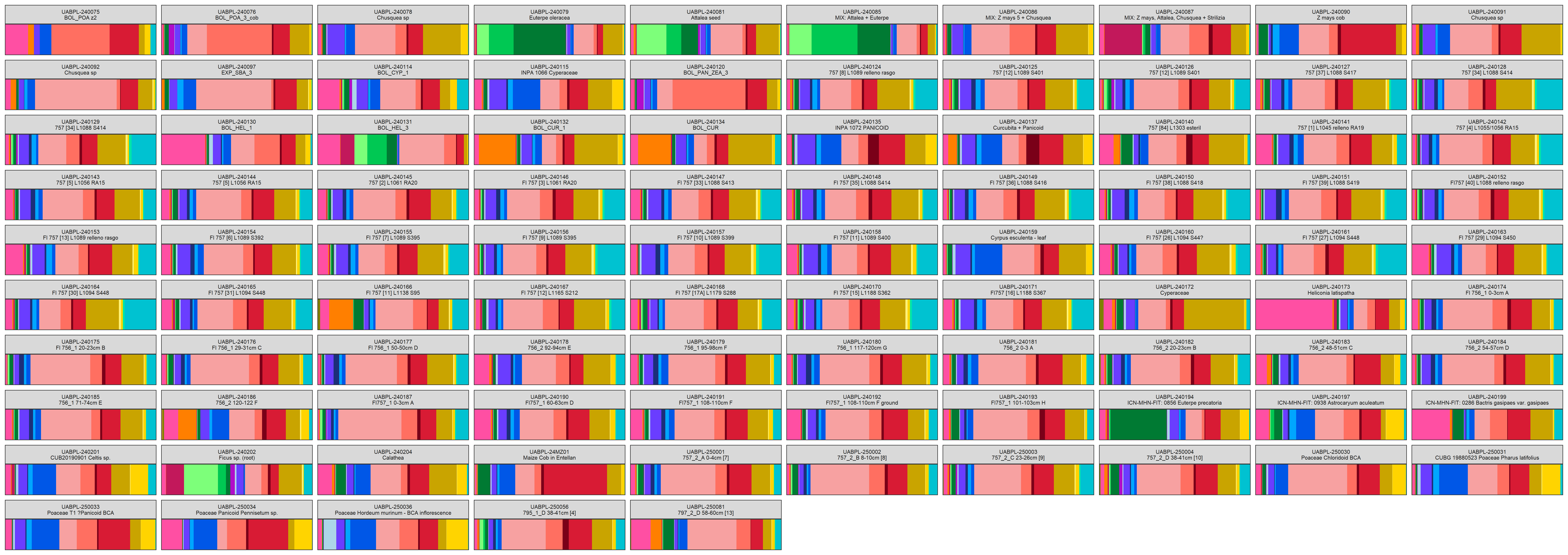}
    \caption{`Barcode' diagram, used to visually the composition of a set of slides. Colours correspond to the same key as Figure \ref{Uncertainty}.}
    \label{Barcode}
\end{sidewaysfigure}

Example uncertainty density plots and barcode diagrams are shown in Figures \ref{Uncertainty} and \ref{Barcode}. The uncertainty density plots summarise the model's confidence for each object's maximum probability prediction, with a greater number of confident predictions indicated by a higher curve farther to the right. Probabilities cannot be below $\frac{1}{n}$, where \textit{n} is the number of possible classes. The position of the sample in compositional space is shown in Figure \ref{Scatterplot}. 

\begin{figure}[t!]
\centering
\includegraphics[width=\linewidth]{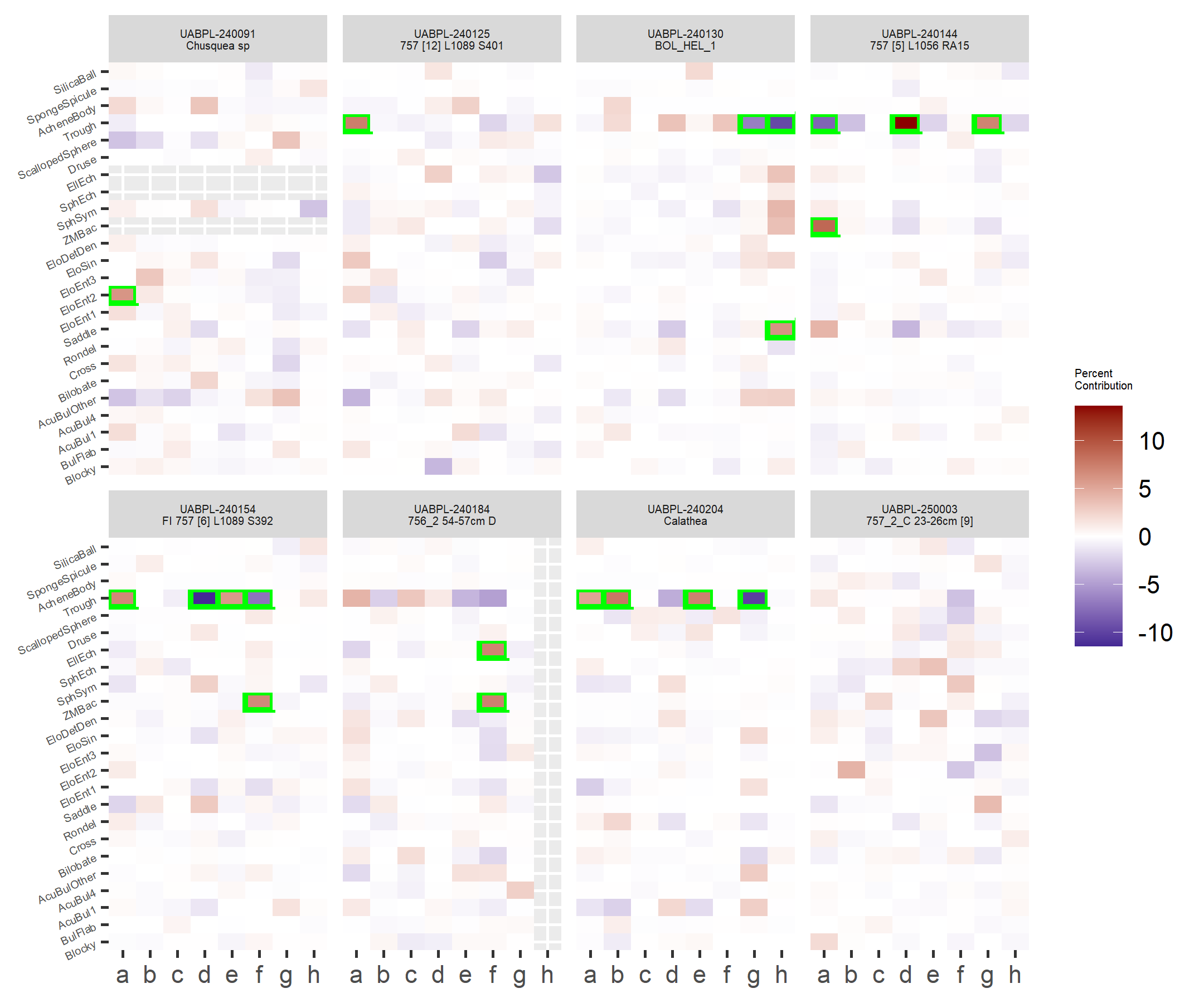}
\caption{Chi square test of independence for slides by sector. Sectors with significant chi square results have highlighted loadings in green.}
\end{figure}

Hierarchical clustering analysis separated samples into two neat clusters, one comprised by samples from reference collection extracts plus sterile archaeological contexts, and the other by the remainder of archaeological contexts (alongside two poorly-digitised maize slides, far left. Figure \ref{Tree}). Within the clear archaeological clustering, there is strong similarity between similar stratigraphic units. Moreover, there is grouping at the family and species level, with maize slides clustering together, as well as palms in their own groups. Finally, the Bayesian modelling was correctly able to identify that prepared mixtures of known components contained palms and Poaceae (Figures \ref{Mixture1} - \ref{Mixture3})

\begin{figure}[t!]
    \centering
    \includegraphics[width=1\linewidth]{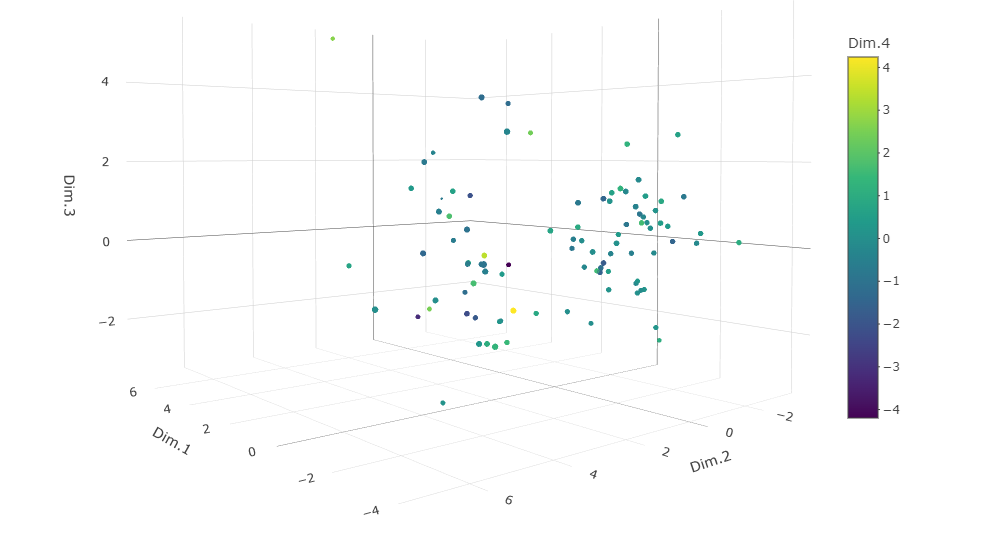}
    \caption{Position of each slide in compositional space according to the predicted morphological compositions.}
    \label{Scatterplot}
\end{figure}

\begin{sidewaysfigure}
\centering
\includegraphics[width=\linewidth]{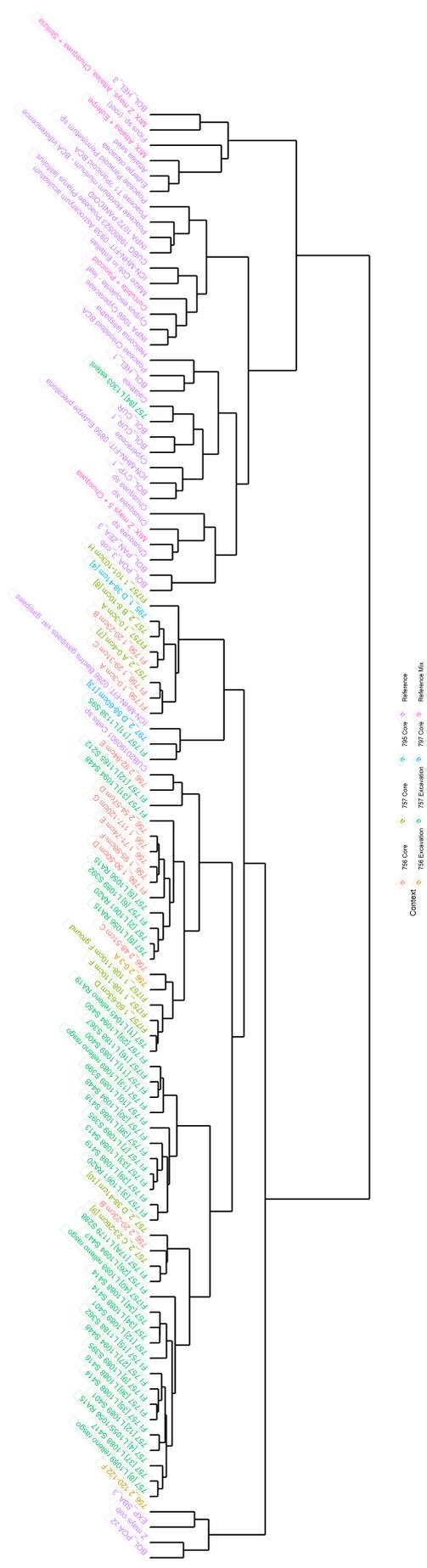}
\caption{Hierarchical clustering analysis of all slides analysed}
\label{Tree}
\end{sidewaysfigure}

\begin{figure}
\centering
\includegraphics[width=\linewidth]{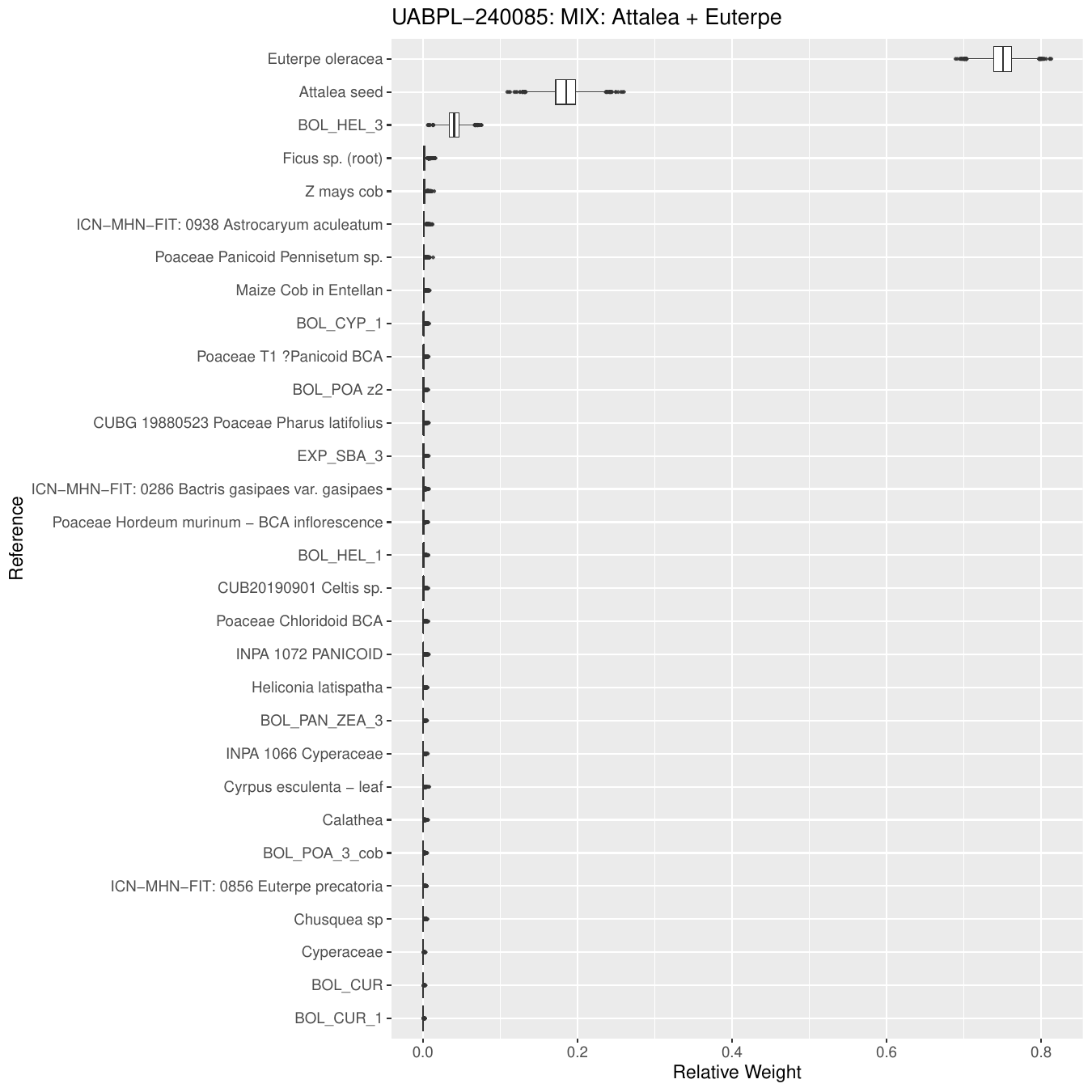}
\caption{Results of the mixture models on a prepared slides of two palm species, \textit{Attalea} sp. and \textit{Euterpe} sp., with the model correctly identifying the correct two palms as the most likely contributors}
\label{Mixture1}
\end{figure}

\begin{figure}
\centering
\includegraphics[width=\linewidth]{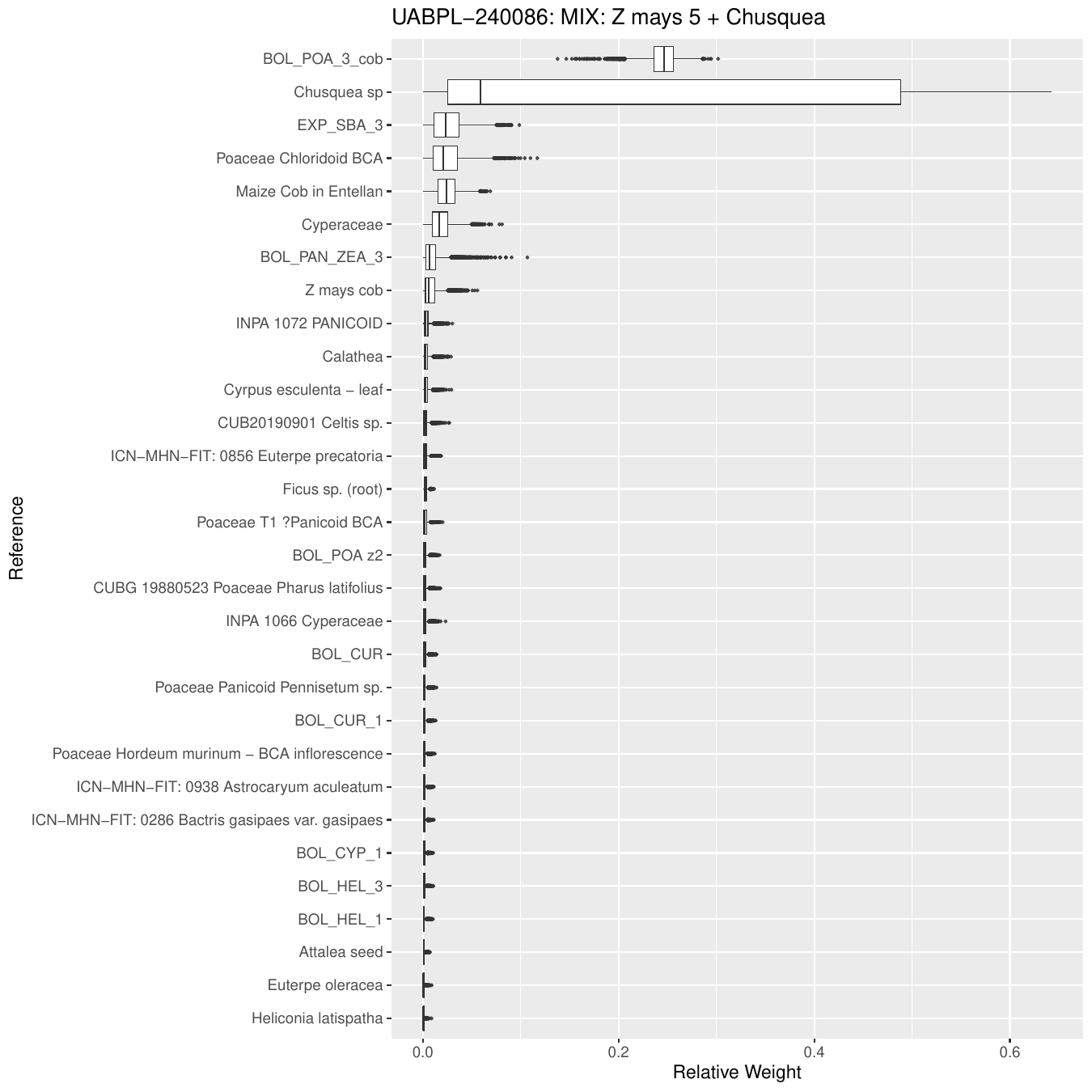}
\caption{Results of the mixture models on a prepared slides of maize and a bamboo species, with the model correctly identifying maize cob and \textit{Chusquea} sp. as the two most probable contributors.}
\label{Mixture2}
\end{figure}

\begin{figure}
\centering
\includegraphics[width=\linewidth]{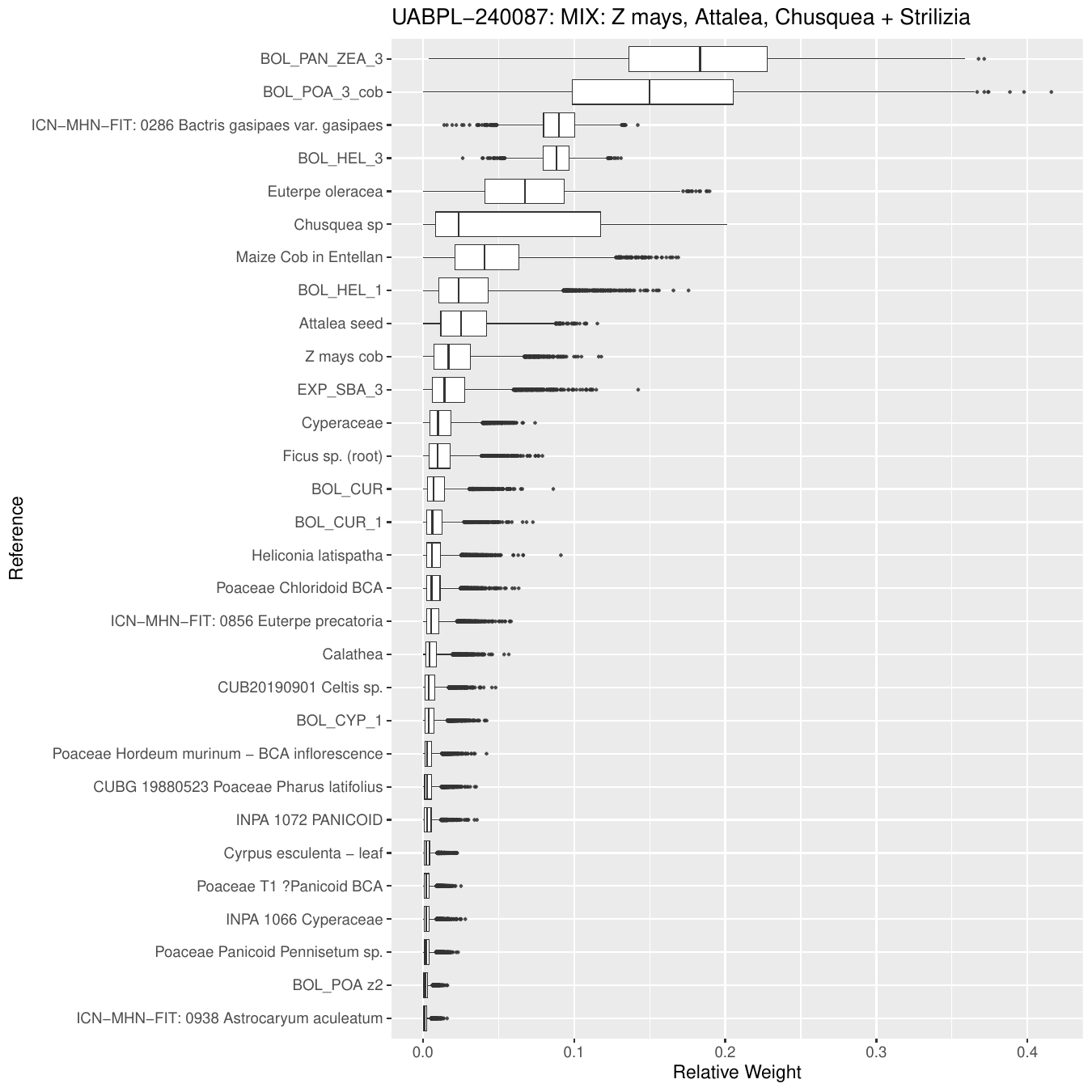}
\caption{Results of the mixture models on a prepared slides of maize, a palm species, a bamboo species, and a plant not in the reference set. Of the six most probable plants identified by the model despite the Strilitzia compound, the first two are maize, three and five are palms, and six is bamboo.}
\label{Mixture3}
\end{figure}

\section{Discussion}

Sorometry demonstrates that phytolith analysis can be reconfigured from a labour-intensive practice centred on small numbers of manually selected objects into a high-throughput, resource-rich analytical framework built around complete digital inventories, probabilistic classification, and assemblage-scale inference. Beyond automating a pre-existing procedure, it treats phytoliths samples as a dense, digitisable population of objects that can be segmented, represented in two and three dimensions, screened for quality, classified, revisited, and analysed as a whole. This shift has implications well beyond efficiency. It moves phytolith research toward an ``omics'' logic in which the analytical object is no longer only the individual morphotype, but the full high-dimensional population of morphologies, conditions, confidences, and co-occurrences that characterise a sample.

This is the sense in which the resources introduced here are as important as the benchmark accuracies. Accuracies for the diagnostic phytoliths included in the present model were all very high. Accuracies for a number of the non-diagnostic morphotypes were generally much lower, but usually still greater than 50\%. Thus, while we are still aways from a perfect classifier, the system employed has shown much success in rapidly surfacing phytoliths that are likely to be a particular class for labelling, and in generating reproducible distributions of predicted phytoliths usable for assemblage analysis. 

Sorometry contributes a complete linked infrastructure for phytolith research: a preprocessing pipeline that converts focus-stacked microscopy into synchronised orthoimages and point clouds; expert-labelled datasets spanning both morphological codes and segmentation-quality classes; hundreds of complete scanned sectors from archaeological and reference slides; pre-trained 2D, 3D, and multimodal models; a graphical environment for expert review and assisted labelling; and downstream tools for compositional analysis and Bayesian source modelling. What is new is therefore not a single classifier, but an integrated resource environment in which data generation, curation, inference, interpretation, and model refinement are all connected. This is precisely what has been missing from most previous AI applications in phytolith research, which were typically limited in taxonomic scope, data scale, dimensionality, and portability to real archaeological assemblages.

A central consequence of this architecture is a dramatic expansion in analytical capacity. Once slides are mounted and scanned, digitisation, segmentation, and classification can proceed continuously and at a pace that is effectively unattainable through conventional microscopy with human analysis. Indeed, according to the model presented in this study, half of the 3.81 million objects from 712 sectors in 123 slides are predicted to be well-segmented phytoliths. In practical terms, this means that phytolith analysis is no longer constrained to the handful of samples or sectors that a specialist can manually inspect within a given project timeline. Entire slide collections can be digitised systematically, revisited without additional microscope time, and classified repeatedly as models improve. This permits denser temporal sequences, broader spatial coverage, and more robust replication across contexts. It also changes project design. Questions that were previously impractical because they required too many counts, too many slides, or too much specialist time become tractable when classification can be applied non-stop to accumulated digital inventories. For archaeology and paleoecology, the result is the possibility of moving from sparse point estimates toward population-scale characterisation of assemblages across stratigraphic, regional, and experimental datasets.

This increase in throughput does not eliminate the importance of expert knowledge; rather, it redistributes it. Expert labour is shifted away from repetitive microscope manipulation and toward higher-value tasks such as defining categories, adjudicating ambiguous forms, curating reference collections, validating edge cases, and interpreting assemblage structure. The GUI is important in this respect because it operationalises a feedback loop between specialists and models. Reviewers are not displaced by automation, but placed in a position to supervise, correct, and strategically extend it. This is especially important in phytolith analysis, where ambiguity, taphonomic alteration, transitional forms, and uneven reference coverage remain intrinsic features of the evidence. The workflow is therefore best understood as an expert-amplification system rather than a replacement for phytolith expertise.

The reproducibility gains are equally significant. Phytolith identification has long been affected by observer subjectivity, uneven access to reference collections, and the difficulty of standardising judgements across analysts and laboratories. Sorometry addresses these problems by making decisions, representations, and processing steps explicit and recoverable. Each segmented object is linked to a persistent digital record; 2D and 3D representations are synchronised; model predictions are stored as probability distributions rather than only hard labels; labelling actions can be reviewed in a consistent interface; and the workflow stages share standardised inputs, defaults, and manifests. In this setting, a classification is not a fleeting visual judgement at a microscope but a revisitable digital event with inspectable provenance. Equally important, the resource includes not only ideal reference forms but also poorly segmented objects, non-phytolith particles, degraded specimens, and ambiguous cases of the kind that generate the greatest disagreement in routine analysis. That broader empirical coverage is essential if reproducibility is to extend beyond curated exemplars to the conditions of real archaeological work.

The results also underscore the scientific value of retaining both 2D and 3D information. Several of the most consequential distinctions in the present study depended on the complementarity of modalities. Image texture and margin decoration carried information that was not always captured in point clouds, whereas three-dimensional form resolved distinctions obscured by orientation in 2D projections. The fusion models therefore matter not just because they improve performance metrics, but also because they more faithfully reflect the epistemic structure of phytolith identification as practiced by specialists, who routinely integrate outline, thickness, relief, surface pattern, and positional context. 

A further strength of the resource is that it supports analysis even when individual predictions are imperfect. Assemblage-scale exploratory tools, clustering, ordination, and Bayesian mixture modelling exploit the fact that systematic prediction structure can remain informative even when some objects are misclassified. This is particularly important for archaeological applications, where the most interesting questions often concern broad compositional differences, latent source mixtures, or the surfacing of unusual samples rather than the unequivocal identification of every individual object. In other words, the value of the system lies not only in finding diagnostic phytoliths, but also in rendering entire assemblages analytically legible. This is one of the most consequential ways in which Sorometry pushes phytolith research toward an ``omics'' paradigm. Besides canonical markers, the signal emerges  from distributions, co-variation, uncertainty structure, and compositional profiles across very large numbers of particles.

The future potential of the platform likely extends beyond the morphotypes currently recognised in the training set. Because Sorometry stores large numbers of segmented objects together with learned feature representations, it creates the conditions for discovering recurrent forms and distributions that may not map neatly onto existing coding schemes. This is especially promising for transitional, degraded, or weakly formalised categories that have historically been difficult to treat consistently. We are already extending the system in this direction through additional CNN architectures, saliency methods for all supervised models, self-supervised representation learning with PointMAE and DINOv2 (\citealt{pang_masked_2022,oquab_dinov2_2024}, and unsupervised classification on CNN-derived feature embeddings. These developments matter for more than incremental performance gains. Saliency can help assess whether models are attending to anatomically meaningful regions; self-supervised learning can leverage the much larger unlabeled corpus than is possible with manual labels alone; and embedding-based clustering may reveal previously underappreciated morphological neighborhoods, candidate diagnostic morphotypes, or characteristic distributional signatures of taxa and contexts. In this sense, the resource is not closed around the categories reported here, but designed to make category formation itself more empirical and extensible.

Recent progress in code-oriented large language models has also materially affected the pace at which such a platform can be built and improved. In our case, the initial pipeline was human-written and comparatively narrow in scope, centred on a more limited model stack and a much less developed interface. Human-supervised use of LLM-assisted software development substantially accelerated the extension of the codebase, making it feasible to implement additional architectures, expand the GUI, integrate experimental modules, and troubleshoot workflow bottlenecks on timescales that would otherwise have been difficult for a small interdisciplinary team. This use of LLMs is methodologically analogous to a shift in much of contemporary industry. There, LLMs increasingly are used to increase engineering capacity and improve software systems without necessarily substituting domain judgement. 

Several challenges remain. Confusion among elongate categories, for example, reflects real morphological continua, degradation, and the inclusion of forms intended to represent archaeological variability rather than only idealised reference exemplars. Future gains will likely come from more targeted augmentation, explicit modelling of degradation states, broader reference collections, and continued refinement of hierarchical or probabilistic category structures. 

Taken together, these results show that Sorometry is not simply a faster way to do conventional phytolith analysis. It is a new research resource that reorganises how phytolith data are generated, standardised, shared, interrogated, and expanded. For phytologists, it provides scalable and reproducible access to richer reference and assemblage information. For archaeologists and paleoecologists, it makes possible forms of sampling intensity and comparative analysis that were previously prohibitive. For AI researchers, it offers a challenging multimodal domain with expert-grounded labels, abundant unlabeled data, and clear opportunities for self-supervision, interpretability, and discovery. What is revolutionary is therefore not only the use of AI to classify phytoliths, but the creation of an extensible analytical ecosystem in which phytolith research can begin to operate at the scale, resolution, and cumulative reproducibility expected of modern data-intensive science.

\section{Conclusion}

The development of Sorometry marks a transformative shift in phytolith research, transitioning the field from a labour-intensive, manual microscopy practice into a high-throughput, ``omics''-scale analytical discipline. By integrating a multimodal AI pipeline that combines ConvNeXt for 2D orthoimages and PointNet++ for 3D point clouds, the system achieves robust classification accuracies of 77.9\% for diagnostic morphotypes and 84.5\% for segmentation quality. Crucially, the inclusion of 3D data proved essential for distinguishing complex morphologies, such as grass silica short cell phytoliths, which are often obscured by their orientation in 2D projections 

Rather than replacing human expertise, Sorometry amplifies it through a purpose-built graphical user interface that facilitates expert review, strategic annotation, and continuous model refinement. Furthermore, by incorporating Bayesian finite mixture modelling, the platform extends analysis from individual object identification to population-level assemblage characterisation, accurately predicting source plant contributions like maize and palms even in complex archaeological mixtures. Ultimately, Sorometry provides a scalable, reproducible analytical ecosystem that standardises expert judgements and enables previously prohibitive, large-scale spatial and temporal analyses in archaeology and paleoecology. 

\newpage
\section*{Acknowledgements}

\noindent This research is an output of the ERC Consolidator project DEMODRIVERS funded by the European Research Council, project no. 101043738. This article contributes to the ICTA-UAB \textit{María de Maeztu Unit of Excellence} (no. CEX2019-000940-M) of the Spanish Ministry of Science, Innovation, and Universities, and also to EarlyFoods SRG (Evolution and Impact of Early Food Production Systems), which has received funding from Agència de Gestió d’\`Ajuts Universitaris i de Recerca de Catalunya (no. SGR-Cat-2021, 00527). We thank J. Iriarte, F. Mayle, G. Morcote-Ríos, the Royal Botanic Garden Edinburg, and the Cambridge University Botanic Garden for providing some of the reference collections used in the AI training.

\clearpage
\newpage

\bibliography{references}
\bibliographystyle{bibstyle.bst}

\newpage

\setcounter{page}{1}

\setcounter{figure}{0}
\setcounter{table}{0}
\setcounter{section}{0}
\renewcommand{\thesection}{S.A\arabic{section}}
\renewcommand{\thetable}{S.A\arabic{table}}
\renewcommand{\thefigure}{S.A\arabic{figure}}

\begin{center}
    {\Large \textbf{Supplemental Information A}}\\
    \vspace*{12pt}
    \textit{\large Glossary of Key Terms}
\end{center}

\input{Glossary_Table}
\newpage

\setcounter{figure}{0}
\setcounter{table}{0}
\setcounter{section}{0}
\renewcommand{\thesection}{S.B\arabic{section}}
\renewcommand{\thetable}{S.B\arabic{table}}
\renewcommand{\thefigure}{S.B\arabic{figure}}

\begin{center}
    {\Large \textbf{Supplemental Information B}}\\
    \vspace*{12pt}
    \textit{\large Phytolith Morphotype Selection and Classification}
\end{center}

Phytoliths were labelled using a quality/object type descriptor (Table \ref{ObjectTypes}), and for well-segmented objects, one of twenty-four morphological labels. The phytolith morphotypes were selected for inclusion in the final training set (Figure \ref{Type_Images}, Tables \ref{Type_Counts} - \ref{Type_Descriptions}. Six of the morphotypes represent phytoliths commonly found in modern and ancient assemblages from diverse plant communities across the globe and have well defined morphological criteria described in ICPN 2.0, despite low diagnostic value. These include \textsc{Acute bulbosus} (split into 3-user defined sub-categories, Table1), \textsc{Blocky}, \textsc{Bulliform flabellate}, \textsc{Elongate entire}, \textsc{Elongate sinuate} and \textsc{Elongate dentate/dendritic}. \textsc{Sponge spicule}s were also included as they exhibit very similar traits to phytoliths (\textsc{Elongate entire}) and are commonly encountered in paleoecological and archaeological assemblages. Four Grass silica short cell phytoliths (GSSCPs); \textsc{Bilobate}s, \textsc{Cross}es, \textsc{Rondel}s and \textsc{Saddle}s were included to encompass a range of diagnostic morphotypes found across different grassland environments. Spheroid Echinate phytoliths, most commonly derived from palms, were sub-categorised into \textsc{Spheroid echinate}, Spheroid symmetrical and \textsc{Ellipsoidal echinate} (Witteveen et al., 2022). Three spheroid phytoliths common in many palm species were included. The palm phytoliths enabled us to test the sensitivity of the AI to differentiate between subtle differences in shape (ellipsoid versus spheroid), number and arrangement of projections. Four other diagnostic phytoliths; \textsc{Druse}s (Zingiberales) Scalloped spheres (\textit{Cucurbita} sp.), \textsc{Trough}s (\textit{Heliconia} sp.) and \textsc{Polygonal achene body} phytoliths (\textit{Cyperaceae}) were included to assess the results on sets of phytoliths with highly distinctive morphologies.

Phytoliths were tagged from a mixture of modern reference and archaeological slides (Table \ref{Provenance}). The aim was to capture the full diversity of each category and conduct the analyses with data which is transferable to archaeological and paleoecological datasets. The reference material enabled us to generate large data sets of target morphologies, while the phytoliths tagged from archaeological slides provided greater diversity, particularly in terms of taphonomy (broken and degraded phytoliths), to prepare the training sets for real world applications.

Some morphotypes consistently performed poorly in the automated classification, which we attribute to the diversity and variety of taphonomic conditions. Elongate categories, for example, included cells with highly degraded margins as well as transitional forms either featuring different margin decoration on different sides, or mixed margin projections. In these cases, phytoliths were labelled according to the dominant user-defined margin decorations, following as closely to examples and descriptions provided in the ICPN 2.0 where possible. Sub-categories of particularly diverse or problematic groups were defined based on morphology (\textsc{Acute bulbosus}; \textsc{Elongate entire}). The morphological criteria for defining each sub-category which was included as a discrete morphotypes in the final training set is summarised in Table \ref{Type_Descriptions}.

\begin{table}
\centering
\caption{Definitions of object types used in classification.}
\label{ObjectTypes}
\begin{tabular}{p{0.28\textwidth} p{0.67\textwidth}}
\hline \hline
\textbf{Object Type} & \textbf{Definition} \\
\hline 
Satisfactory phytolith & Single, well-segmented phytoliths or other objects where all or most key morphological features are captured in the point cloud and segmentation. Some low-level additional detritus or noise is acceptable. \\

Fragmentary phytolith & Only part of the phytolith is captured in the point cloud and segmentation, or part of the phytolith falls out of the field of view. Some diagnostic features are still captured. \\

Poorly-segmented phytoliths & More than one discrete object or phytolith is captured in the segmentation. \\

Multicell & Two or more phytoliths joined in anatomical connection, or phytoliths with other parts of the silica structure attached. \\

Trash & Segmentation detritus; objects capturing only undiagnostic elements of a whole phytolith (e.g.\ only one edge); small non-phytolith objects ($< 5\,\mathrm{\mu}$m); other non-phytolith material (e.g.\ quartz, clay aggregates); silica fragments; and segmented objects with such low image and point-cloud resolution that they cannot be securely identified as phytoliths. \\

Sponge & \textsc{Sponge spicule}. \\

Diatom & Whole diatom or fragment of diatom. \\
\hline \hline
\end{tabular}
\end{table}

\begin{table}[b!]
\centering
\caption{Segmentation Quality and Object Type Count (Train + Test)}
\label{AllClasses-Count}
\begin{tabular}{lr}
\hline \hline
\textbf{Group} & \textbf{Total Count} \\
\hline
Multicell & 281 \\
PoorlySegmented & 3060 \\
Singlet & 8859 \\
\textsc{SpongeSpicule} & 68 \\
Trash & 771 \\
\hline \hline
\end{tabular}
\end{table}

\begin{table}[b!]
\centering
\caption{Segmentation Quality and Object Type UABPL-240151d (Validation Set)}
\begin{tabular}{lr}
\hline
Group & Total Count \\
\hline
Multicell & 8 \\
PoorlySegmented & 260 \\
Singlet & 1402 \\
\textsc{SpongeSpicule} & 9 \\
Trash & 521 \\
\hline
\end{tabular}
\end{table}

\begin{table}
\centering
\caption{Morphotype Count (Train + Test)}
\label{Type_Counts}
\begin{tabular}{lr}
\hline \hline
\textbf{Group} & \textbf{Total Count} \\
\hline
\textsc{AcheneBody} & 17 \\
\textsc{AcuBul1} & 174 \\
\textsc{AcuBul4} & 116 \\
\textsc{AcuBulOther} & 183 \\
\textsc{Bilobate} & 344 \\
\textsc{Blocky} & 301 \\
\textsc{BulFlab} & 143 \\
\textsc{Cross} & 155 \\
\textsc{Druse} & 84 \\
\textsc{EllEch} & 68 \\
\textsc{EloDetDen} & 186 \\
\textsc{EloEnt1} & 92 \\
\textsc{EloEnt2} & 128 \\
\textsc{EloEnt3} & 106 \\
\textsc{EloSin} & 184 \\
\textsc{Rondel} & 422 \\
\textsc{Saddle} & 411 \\
\textsc{ScallopedSphere} & 432 \\
\textsc{SilicaBall} & 261 \\
\textsc{SphEch} & 128 \\
\textsc{SphSym} & 74 \\
\textsc{SpongeSpicule} & 66 \\
\textsc{Trough} & 198 \\
\textsc{ZMBac} & 93 \\
\hline \hline
\end{tabular}
\end{table}

\begin{table}[htbp]
\centering
\caption{Morphotype Validation Set from UABPL-240151d}
\begin{tabular}{lr}
\hline \hline
\textbf{Group} & \textbf{Total Count} \\
\hline
\textsc{AcuBul4} & 10 \\
\textsc{AcuBulOther} & 28 \\
\textsc{Bilobate} & 15 \\
\textsc{Blocky} & 51 \\
\textsc{BulFlab} & 12 \\
\textsc{Cross} & 3 \\
\textsc{EloDetDen} & 11 \\
\textsc{EloEnt1} & 2 \\
\textsc{EloEnt2} & 7 \\
\textsc{EloEnt3} & 33 \\
\textsc{EloSin} & 31 \\
\textsc{Rondel} & 22 \\
\textsc{Saddle} & 52 \\
\textsc{SpongeSpicule} & 9 \\
Outside Training Set & 1131 \\
\hline \hline
\end{tabular}
\end{table}

\begin{landscape}
\begin{longtable}{p{0.11\linewidth}p{0.25\linewidth}p{0.15\linewidth}p{0.45\linewidth}}
\caption{Key phytolith morphotype groups included in the training set, with corresponding names used in the confusion matrix. Descriptions are based on ICPN 2.0 unless otherwise stated.}
\label{Type_Descriptions}\\
\toprule
Code & Name & Name in confusion matrix & Key morphological attributes \\
\midrule
\endfirsthead

\toprule
Code & Name & Name in confusion matrix & Key morphological attributes \\
\midrule
\endhead

\midrule
\multicolumn{4}{r}{\emph{Continued on next page}}\\
\midrule
\endfoot

\bottomrule
\endlastfoot

ACUBUL & \textsc{Acute bulbosus} & \textsc{AcuBul1}; \par \textsc{AcuBul4}; \par \textsc{AcuBulOther} & Narrow body with acute apex and wider antapex. Size between $\sim$25--100~$\mu$m \\
BLO & \textsc{Blocky} & \textsc{Blocky} & Solid 3D bodies with length/width $< 2$ \\
BULFLA & \textsc{Bulliform}\textsc{flabellate} & \textsc{BulFlab} & Well silicified ``fan shape'', lower part narrower than upper part \\
ELOENT & \textsc{Elongate entire} & \textsc{EloEnt1}; \textsc{EloEnt2}; \textsc{EloEnt3} & Rectilinear 2D outline with predominately entire (smooth) margins. L:W $> 2$ \\
ELOSIN & \textsc{Elongate sinuate} & \textsc{EloSin} & Rectilinear 2D outline with sinuate margins (alternating concavities and convexities). L:W $> 2$ \\
ELODET/DEN & \textsc{Elongate dentate/dendritic} & \textsc{EloDetDen} & Rectilinear 2D outline with dentate margins (acute projections), and occasionally branched processes. L:W $> 2$ \\
GSSCP - BIL & \textsc{Bilobate} & \textsc{Bilobate} & OPS consists of 2 lobes separated by 2 indentations or a distinct castula, with length $\geq$ 1:3 the width of lobes \\
GSSCP - CRO & \textsc{Cross} & \textsc{Cross} & 4 roughly equal lobes in OPS separated by indentations. Approximately equal width:length in planar view. Longest dimension $< 1$:3 times the length of the dimension at right angles to it \\
GSSCP - RON & \textsc{Rondel} & \textsc{Rondel} & Circular to oval OPS (with indentations of flattened on 1 aspect). IPS variable rounded to angular, pointed or carinate). \\
GSSCP - SAD & \textsc{Saddle} & \textsc{Saddle} & Symmetrical with 2 convex faces connected by concave faces \\
ZMBAC & \textsc{Tabular tuberculate} & \textsc{ZMBac} & Elongate, rectangular or irregular in 2D, thin (tabular) in 3D with distinctive spines on all margins (dentate to dendritic), and surface texture psilate, often with irregular bacculate-tuberculate-echinate protrusions \\
POLACH & \textsc{Polygonal achene body} & \textsc{AcheneBody} & 4--8 generally straight edges, with dense to light granulate (stippled) surface decoration \\
SSP & \textsc{Scalloped Sphere} & \textsc{ScallopedSphere} & Large spherical forms with scalloped surfaces of deeply contiguous concavities (Bozarth, 1987, p.~608) \\
TRO & \textsc{Trough} & \textsc{Trough} & Deep, centrally located troughs (cavate), with decorated or smooth surfaces (Piperno, 2006, p.~38) \\
SPHECH & \textsc{Spheroid echinate} & \textsc{SphEch} & Spheroidal 3D shape with echinate texture, 10--12 surficial projections \\
SPHSYM & \textsc{Spheroid echinate symmetrical} & \textsc{SphSym} & Spheroidal 3D shape with echinate texture, 8--10 surficial projections symmetrically arranged \\
ELLECH & \textsc{Ellipsoidal echinate} & \textsc{EllEch} & Ellipsoidal 3D shape with echinate texture, 10--12 surficial projections \\
DRU & \textsc{Druse} & \textsc{Druse} & Irregularly spheroidal 3D with irregular echinate/spiny surface texture, sometimes appearing ``folded'' \\
SPO & \textsc{Sponge spicule} & \textsc{SpongeSpicule} & NP: Rectilinear elongate outline with entire margins and axial canal \\
SB & \textsc{Silica ball} & \textsc{SilicaBall} & NP: Spherical with smooth (psilate) surface and margins, 20~$\mu$m diameter \\

\end{longtable}
\noindent Notes: OPS --- Outer periclinal surface/aspect; IPS --- Inner periclinal surface/aspect; NP --- non-phytolith. Descriptions are based on ICPN 2.0 unless otherwise stated.
\end{landscape}

\begin{table}[b!]
\footnotesize
\caption{Provenance of scanned microscopic objects used in this study.}
\label{Provenance}
\begin{longtable}{l c c c}
\hline

\textbf{} & \textbf{Slide \&} & \textbf{} & \textbf{Count} \\
\textbf{Context} & \textbf{Sector ID} & \textbf{Type} & \textbf{(Morphotype/Quality)} \\
\hline
\multirow[t]{3}{*}{FI 757 [10] L1089 S399} & UABPL-240157a & ARCH & 232 / 1405 \\
 & UABPL-240157b & ARCH & 208 / 1041 \\
 & UABPL-240157c & ARCH & 311 / 869 \\
FI 757 [27] L1094 S448 & UABPL-240161a & ARCH & 1 / 35 \\
FI 757 [39] L1088 S419 & UABPL-240151b & ARCH & 3 / 4 \\
FI 756\_1 0-3cm A & UABPL-240174a & CORE & 193 / 1437 \\
FI 756\_1 50-50cm D & UABPL-240177a & CORE & 210 / 1001 \\
FI 756\_1 117-120cm G & UABPL-240180h & CORE & 156 / 520 \\
FI 756\_1 95-98cm F & UABPL-240179a & CORE & 182 / 1057 \\
FI 756\_2 0-3 A & UABPL-240181a & CORE & 2 / 7 \\
FI 756\_2 120-122 F & UABPL-240186a & CORE & 3 / 6 \\
FI 757\_1 101-103cm H & UABPL-240193b & CORE & 277 / 395 \\
FI 757\_2 23-26cm C [9] & UABPL-250003a & CORE & 2 / 4 \\
FI 795\_1 38-41cm D [4] & UABPL-250056a & CORE & 4 / 12 \\
\multirow[t]{6}{*}{\textit{Cucurbita} sp.} & UABPL-240132a & REF & 71 / 89 \\
 & UABPL-240132b & REF & 79 / 101 \\
 & UABPL-240132c & REF & 91 / 122 \\
 & UABPL-240132d & REF & 73 / 117 \\
 & UABPL-240132e & REF & 39 / 67 \\
 & UABPL-240132f & REF & 81 / 121 \\
\textit{Heliconia} sp. & UABPL-240130a & REF & 184 / 369 \\
\multirow[t]{3}{*}{\textit{Zea mays}} & UABPL-240120a & REF & 245 / 550 \\
 & UABPL-240120b & REF & 109 / 251 \\
 & UABPL-240120c & REF & 59 / 95 \\
\multirow[t]{2}{*}{\textit{Chusquea} sp.} & UABPL-240078a & REF & 9 / 57 \\
 & UABPL-240078e & REF & 1 / 1 \\
\multirow[t]{5}{*}{Cyperaceae} & UABPL-240172a & REF & 6 / 16 \\
 & UABPL-240172b & REF & 4 / 6 \\
 & UABPL-240172d & REF & 1 / 6 \\
 & UABPL-240172e & REF & 5 / 21 \\
 & UABPL-240172h & REF & 2 / 5 \\
\multirow[t]{2}{*}{\textit{Heliconia latispatha}} & UABPL-240173a & REF & 40 / 151 \\
 & UABPL-240173b & REF & 8 / 31 \\
\textit{Euterpe precatoria} & UABPL-240194a & REF & 32 / 222 \\
Panicoid & UABPL-240135a & REF & 284 / 677 \\
 \textit{Attalea}  + \textit{Euterpe} & UABPL-240085d & REF & 234 / 292 \\
 \textit{Zea mays} + \textit{Chusquea} sp. & UABPL-240086b & REF & 3 / 103 \\
 \textit{Zea mays},\textit{ Attalea }sp., \textit{Chusquea} sp. + Strilizia & UABPL-240087a & REF & 88 / 95 \\
\multirow[t]{3}{*}{Poaceae Chloridoid BCA} & UABPL-250030a & REF & 245 / 434 \\
 & UABPL-250030b & REF & 9 / 45 \\
 & UABPL-250030d & REF & 19 / 37 \\
Poaceae \textit{Hordeum murinum} - BCA inflorescence & UABPL-250036a & REF & 299 / 414 \\
\multirow[t]{3}{*}{Poaceae Panicoid \textit{Pennisetum} sp.} & UABPL-250034a & REF & 1 / 8 \\
 & UABPL-250034b & REF & 40 / 55 \\
 & UABPL-250034d & REF & 4 / 5 \\
Poaceae T1 Panicoid BCA & UABPL-250033b & REF & 217 / 401 \\
\hline
\end{longtable}
\end{table}

\clearpage
\newpage

\setcounter{figure}{0}
\setcounter{table}{0}
\setcounter{section}{0}
\renewcommand{\thesection}{S.C\arabic{section}}
\renewcommand{\thetable}{S.C\arabic{table}}
\renewcommand{\thefigure}{S.C\arabic{figure}}

\begin{center}
    {\Large \textbf{Supplemental Information C}}\\
    \vspace*{12pt}
    \textit{\large Workflow Methodology}
\end{center}

The implementation emphasis throughout Sorometry is throughput and reproducibility under realistic laboratory constraints rather than optimisation for a single benchmark metric. Practically, this means we preferred methods that are robust to variable slide quality, permit batch processing of very large inventories, preserve provenance through run manifests, and can be operated by domain specialists through a graphical interface rather than requiring repeated ad hoc scripting.

The software stack is organised as scriptable workflow stages with shared configuration and synchronised defaults. Root-level Python wrappers call stage-specific scripts, write manifests, and expose parameters in a way that is mirrored in the GUI so command-line and GUI runs remain methodologically equivalent. R scripts are retained for assemblage compositional analysis and Bayesian mixture modelling because they offer mature implementations of compositional transforms, multivariate ordination, clustering, and Stan-based posterior inference with minimal reinvention.

\section{Slide Scanning}
Slides were digitised on an Olympus VS200 brightfield platform using 185 multi-plane focus stacks (z-slices) captured in two chunks sampled to capture the lowermost 49 \textmu{}m of mounting depth, where settled phytoliths are concentrated under the laboratory mounting protocol described in the main text. We employed an Olympus UPLXAPO60XO objective which employs immersion fluid and a maximum image resolution of 0.091 $\mathrm{\mu}$m per pixel. Scanning was organised into 2 mm × 2 mm sectors, producing per-sector image stacks with enough axial sampling to permit focus-based depth recovery. The system's VSI file outputs were converted to JPG using the \textit{VSI Convert }plugging, with the minimum compression possible enabled.

Digital microscope JPG name outputs were standardised with\textit{ Image\_Convert.py}, which remaps Olympus export naming into a stable slide-sector convention and moves image planes and metadata into standardised directories used by downstream scripts. The conversion logic parses section and z-slice identifiers from Olympus filenames, replaces scanner-specific identifiers with harmonised UABPL slide number (for \textbf{\textit{U}}\textit{niversitat}\textit{ }\textbf{\textit{A}}\textit{utònoma}\textit{ de }\textbf{\textit{B}}\textit{arcelona}\textit{ }\textbf{\textit{P}}\textit{hyto}\textbf{\textit{l}}\textit{ith})\textit{ }slide IDs linked to the master slide inventory, preserves per-sector image order, and propagates both \textit{.}\textit{ini} and XML metadata files needed for physical scale recovery.

\section{Pointcloud and Orthoimage Generation}

The point-cloud/orthoimage stage (\textit{Point\_Extraction.py}) converts each z-stack into two synchronised representations: a BL15 orthomosaic image and an XYZRGB point cloud. For each z-slice, an edge-preserving bilateral filter is applied (diameter \textit{bilateral\_d} = 15,  and colour sensitivity \textit{sigmaColor} = 35), followed by Laplacian edge response detection (\textit{ksize} = 3) as a focus proxy. At each x-y location, the z-slice with maximal Laplacian response supplies the retained RGB value in the orthomosaic. Candidate 3D points are then thresholded by a log-space criterion defined as  mean(log(\textit{L }+ 1 )), where \textit{L }is the value of the pixel plus a configurable multiple of the log-space standard deviation (\textit{filter\_value\_default}\textit{ }= 3.75, with slide-specific override \textit{filter\_value\_white} = 4.0 for known high-brightness slides). This thresholding strategy intentionally biases toward high-confidence, edge-rich voxels for segmentation robustness and compute efficiency at scale. Final orthomosaics are smoothed with a larger bilateral diameter (\textit{final\_image\_bilateral}\_d = 20) to stabilise visualisation and crop rendering without modifying the already exported point cloud.

A specific correction path is implemented for stacks scanned in two chunks around common zero-focus planes. In these cases, we extract the middle ninth of the image corresponding to the zero-focus plane in each stack. We estimate the chunk offset by applying a Fast Fourier Transform-based correlation analysis on the two crops, taking the coordinates of the pixel of highest value as the tile offset. Inter-chunk colour gain differences are estimated using a linear regression between each individual colour channel in the common zero-slice, propagating all changes across the top chunk. This avoids seam and mirror effect artifacts in orthomosaics and RGB point attributes.

\section{Object Segmentation}
Individual object candidates are segmented in \textit{Initial\_Segmentation.py}\textit{ } using CloudComPy connected-component extraction on full-sector point clouds. Before segmentation, index-space coordinates are converted back to metric space using scan metadata scale factors from the slide INI files, and RGB values are attached to the CloudComPy point cloud object. Connected components are extracted with \textit{octree\_level} = 12, and \textit{max\_number\_components} = 99999. This octree-based strategy was selected over more computationally expensive density clustering alternatives because it scales well to our corpus with millions of objects---often times tens of thousands per sector---and provides deterministic partitions tied to spatial adjacency in 3D. The \textit{min\_component\_size} = 750 was set high enough to suppress very small particulate noise while retaining fragmented but still classifiable phytolith forms.

Each retained component is exported as a compressed segment cloud with integer xyzrgb columns, after reprojecting metric coordinates back into index space for compact storage and compatibility with existing model loaders. The orthographic image crop for each segment is generated downstream by applying each segment's projected bounding box to the sector-level orthoimage, ensuring that 2D and 3D representations remain geometrically linked at the sample ID level. This one-cloud/one-crop pairing is critical for late fusion modelling and for GUI-based expert review where analysts assess both projection texture and 3D shape cues.

\section{GUI-based Expert Review}
Human annotation and review are performed in the PyQt-based Sorometry GUI (\textit{sorometry\_}\textit{gui}\textit{.py}), which is intentionally designed for rapid serial review of large segment sets while preserving per-decision provenance. For each sample, reviewers can inspect the true-scale image crop with a toggleable projected cloud overlay, navigate among sectors and sample IDs, and open 3D representations in Open3D for manual rotation when orientation-dependent morphology is ambiguous in 2D. Label records are persisted with required reviewer initials, timestamp fields, segmentation-quality/value assignments, free-text notes, and morphology code entries. The class table schema includes both quality labels and morphotype codes, enabling a two-stage supervisory regime in which segment quality filtering and morphology prediction are trained separately. To avoid burnout, the GUI allows the user to switch between tagging modes, allowing them to focus effort refining particular tags through label and probability filtering in single-object and gallery views instead of a pre-defined structure.

The GUI can ingest probability tables from previously trained models and render ranked predictions for the active sample. Reviewers can accept suggested classes directly, inspect full probability vectors, and map predictions to codebook entries when a one-to-many relationship exists between machine class names and local coding conventions. Prediction generation can be triggered for either full slides or individual phytolith IDs, allowing targeted update of uncertain records without rerunning complete inference passes. This design was chosen to optimise reviewer time and labelling throughput. Full-manual coding remains  paramount for  ambiguous specimens, but model-prioritised and model-assisted navigation substantially reduces interaction costs in high-volume curation rounds and surfaces objects with morphotypes currently absent from the labelled corpus.

\section{Training and Dataset Generation}
Training sets are assembled from reviewer-labelled exports using \textit{Pointcloud\_Formatting.p}\textit{y} (morphological predictions) and \textit{Pointcloud\_Formatting\_All.py} (all-class quality/particle-type sets). Eligible rows are filtered by configurable quality values and tag exclusions, grouped by code-reviewer combinations (or class-reviewer combinations for all-class runs), and split into train/test partitions using stratified per-class tokenisation with a default \textit{train\_ratio}\textit{=0.8}. We name datasets and associated models based on $<$LabelType$>$-$<$DateDataCompiled$>$-$<$RandomWord$>$, with the Label Types being either \textit{Classified} for morphological codes, and \textit{AllClasses} for quality and object type classes. In the walnut-associated all-class run (\textit{AllClasses-20260220-walnut}) quality, we employed 10,431 training instances and 2,608 test instances, with classes spanning Singlet, PoorlySegmented, Trash, Multicell, and \textsc{SpongeSpicule}, whereas for the morphological run (\textit{Classified-20260220-walnut) }records 3,531 training instances and 883 test instances across 24 morphological classes (Table \ref{Type_Counts}). This split supports the workflow's quality-gating stage by providing robust negative and low-quality categories before morphology-specific inference is applied.

Three supervised model families are used in the reported experiments---one relying on 2D information, one on 3D information, and one on both. The first is ConvNeXt (\citealt{liu_convnet_2022}) for image-only classification (\textit{Train\_ConvNeXt.py}), using \textit{convnext\_tiny} architecture, image size 384 pixels, AdamW optimisation (\textit{learning\_rate} = 3 x 10-4, \textit{weight\_decay} = 0.05), cosine scheduling with warmup, label smoothing (0.1), and moderate MixUp (0.2, \textit{cutmix} = 0). Augmentations were microscopy-tailored and intentionally mild (random resized crop around native object framing, horizontal and vertical flips, limited rotation, modest brightness/contrast/saturation perturbation, and mild blur/sharpness controls) to preserve diagnostic fine structure while still reducing overfitting.

The second is PointNet++ for cloud-only classification (\textit{Train\_Pointnet2.py}, although the Sorometry also supports PointNeXt and Pointnet) (\citealt{Qi2017,qian_pointnext_2022}) , with the FocusTomography fork of the PointNet pipeline configured for raw cloud loading and metadata-aware scaling; representative walnut runs used \textit{pointnet2\_cls\_ssg} architecture with an added late-fusion scalar corresponding to log10\textit{N} (where \textit{N }is the number of points---to make the model scale-aware),  \textit{num\_points}\textit{ }= 4096, Adam optimisation (\textit{learning\_rate}=10-3), and long-horizon training due to model complexity (epochs = 200).

The third is a multimodal fusion model. Fusion training (\textit{Train\_Fusion.py}) combined a PointNet++ encoder (with the late-fusion scalar aforedescribed) and ConvNeXt encoder with a transformer-style cross-modal module (\textit{fusion\_layers} = 2\textit{, }\textit{fusion\_dim} = 256, \textit{fusion\_heads}=4). The classifier blends the point-only, image-only, and fused-token logit streams, with learned gates that can reweight modalities when one branch is degraded. Supervised fusion optimisation used 100 epoch schedules in recorded runs, \textit{learning\_rate}\textit{ }= 3 x 10-4, \textit{weight\_decay} = 0.05), warm-up of 5 epochs, label smoothing (0.1), MixUp (0.2), and AMP.

\section{Assemblage Statistics}
Assemblage-scale analysis is implemented in\textit{ }\textit{Slide\_Assemblage.R}. For each slide-sector, the script reads model probability tables, assigns maximum-probability classes, merges with all-classes screening output, and removes objects predicted as poorly segmented, trash, or multicell before morphology summaries. Per-slide outputs include stacked composition plots and class-wise confidence density plots.

Because phytolith class frequencies are compositional, analyses are conducted on isometric log-ratio transformed proportions rather than raw counts in Euclidean space. Principal components analysis on ILR-transformed data is used for low-dimensional structure discovery at sector and slide levels, and Ward-linkage agglomerative clustering (\textit{cluster::}\textit{agnes}, employing Ward's  clustering method) is used for dendrogram-based grouping. Sector representativeness within slides is assessed by chi-square tests of independence on sector-by-class contingency tables after dropping all-zero classes per slide. Contribution heatmaps are generated from signed percent contributions of standardised residuals retaining residual sign, with visual highlighting for high-contribution cells in statistically significant slides.

\section{Assemblage Modelling}
Mixture analysis is implemented in \textit{Mixture\_Model.R} and wrapped by \textit{Run\_Mixture\_Model.py}. The model is a Bayesian finite mixture of multinomials coded in Stan, where each reference slide defines one process with class-probability vector $\mathbf{\pi}_i$, test slide counts $\omega$ are multinomial with implied probabilities $\mathbf{q} = \mathbf{\Pi}^\mathrm{T}\mathbf{\omega}$, and mixture weights  follow a Dirichlet prior (default weakly informative \textit{alpha} = 1 for all components). The script defensively renormalises each reference process row to sum to one before inference. Posterior inference uses Hamiltonian Monte Carlo through \textit{rstan}\textit{::sampling} with default settings of 4 chains, 2,000 iterations per chain, 1,000 warmup, \textit{adapt\_delta} = 0.9, and \textit{max\_treedepth} = 12; in the provided looped application, each test sample is fit with 7 chains to stabilise posterior summaries for report plots and ensure that not only the maximum peak is explored in possible multimodal data.

Posterior outputs include weight draws, implied class-probability draws, and posterior predictive replicated counts. Results are exported as per-slide boxplot PDFs and CSV tables of posterior weight draws. This model was favoured over distance-only source attribution methods because it preserves count-scale likelihood structure and yields directly interpretable uncertainty on source contributions. This allows for a sample-level inference on the original composition of plants that contributed to the sample. Relative to non-Bayesian EM mixtures, the Stan implementation is computationally heavier but materially more informative for low-abundance contributors and near-collinear reference compositions, both of which are common in archaeobotanical assemblages.

\newpage
\setcounter{figure}{0}
\setcounter{table}{0}
\setcounter{section}{0}
\renewcommand{\thesection}{S.D\arabic{section}}
\renewcommand{\thetable}{S.D\arabic{table}}
\renewcommand{\thefigure}{S.D\arabic{figure}}

\begin{center}
    {\Large \textbf{Supplemental Information D}}\\
    \vspace*{12pt}
    \textit{\large Additional Confusion Matrices}
\end{center}

\input{Confusion_Supplement}

\newpage
\setcounter{figure}{0}
\setcounter{table}{0}
\setcounter{section}{0}
\renewcommand{\thesection}{S.E\arabic{section}}
\renewcommand{\thetable}{S.E\arabic{table}}
\renewcommand{\thefigure}{S.E\arabic{figure}}

\begin{center}
    {\Large \textbf{Supplemental Information E}}\\
    \vspace*{12pt}
    \textit{\large Learning curves for classification models}
\end{center}

\begin{figure}[b!]
    \centering
    \includegraphics[width=\linewidth]{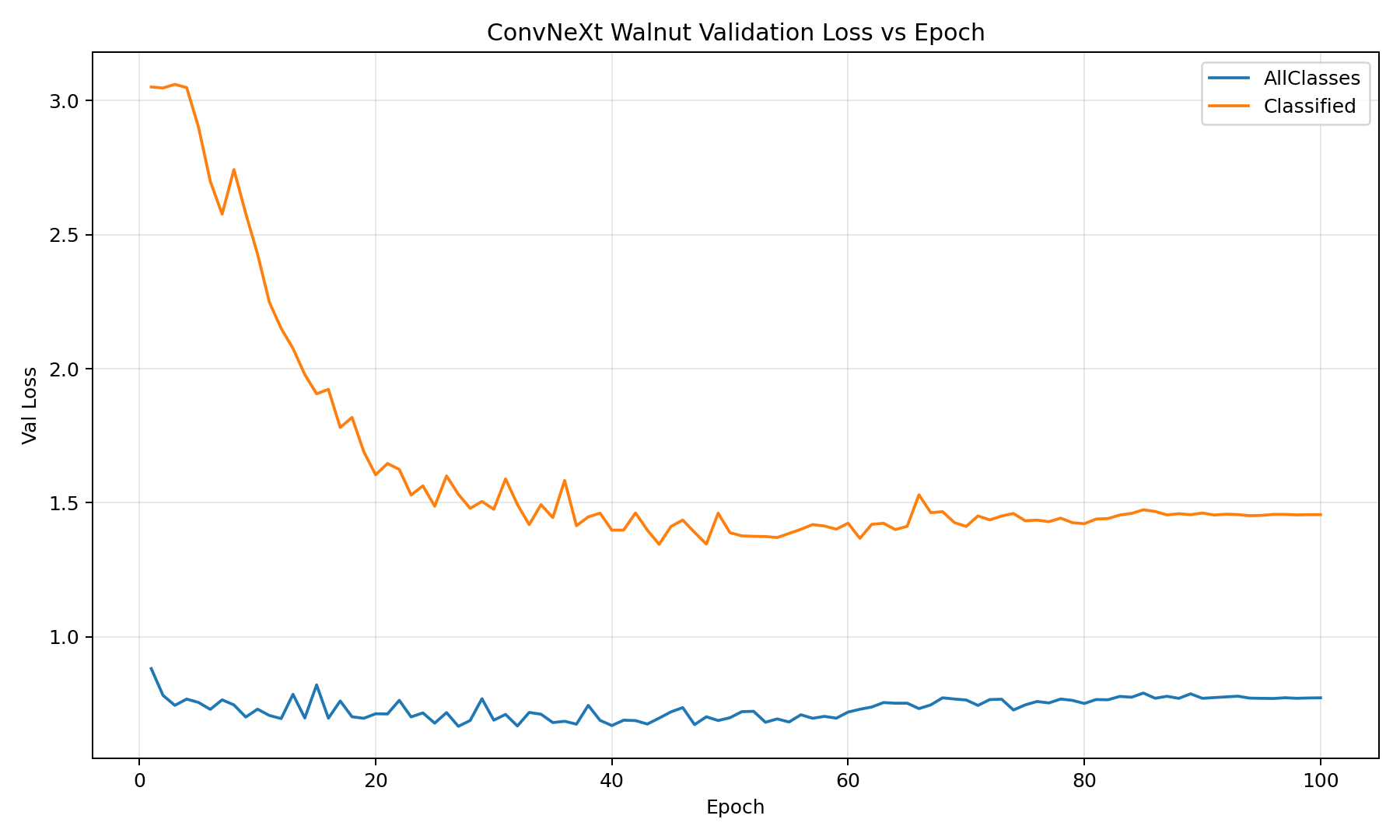}
    \caption{Learning Curve for the ConvNeXt 2D image morphological model (Classified) and segmentation quality/object type model (AllClasses).}
\end{figure}

\begin{figure}
    \centering
    \includegraphics[width=\linewidth]{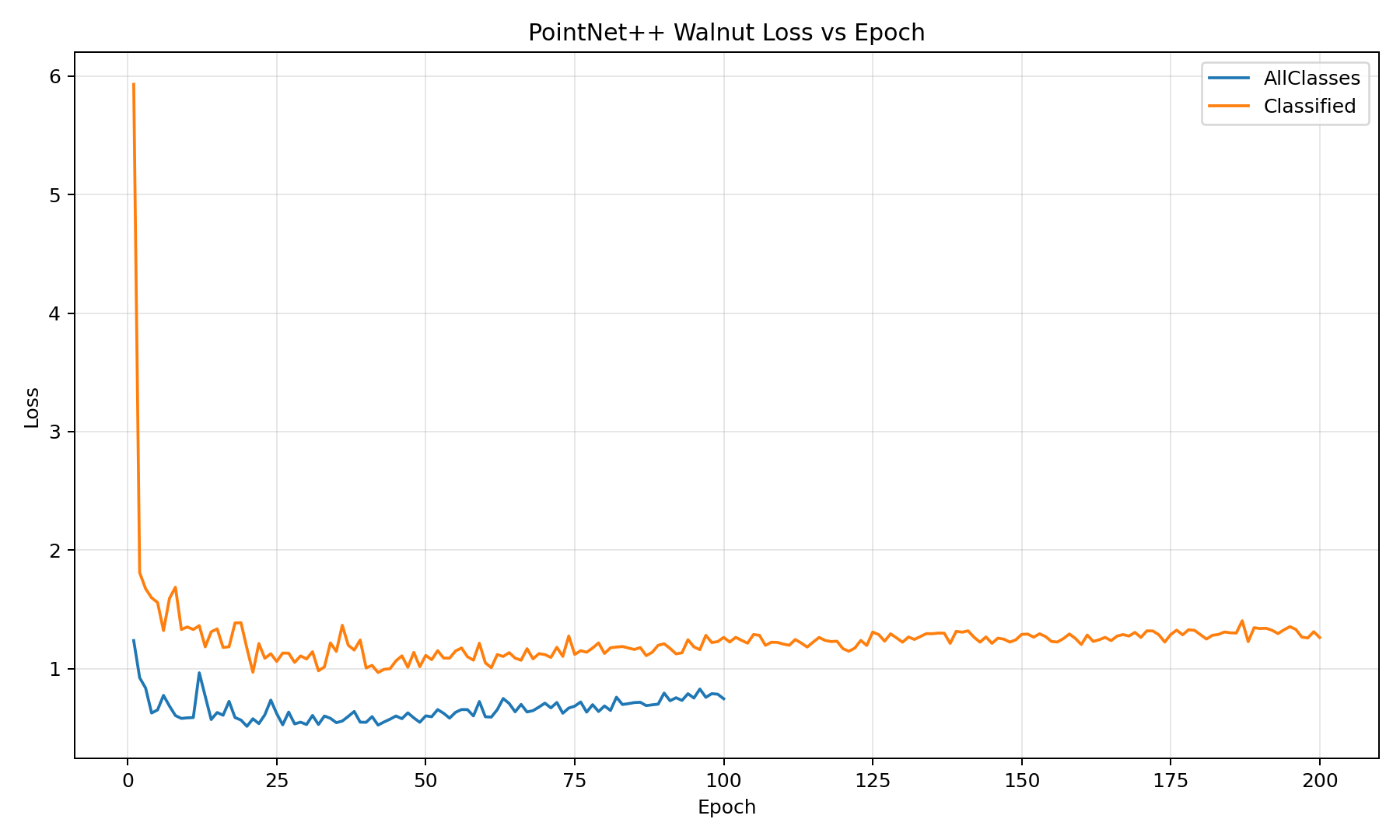}
    \caption{Learning Curve for the Pointnet++ 3D pointcloud  morphological model (Classified) and segmentation quality/object type model (AllClasses).}
\end{figure}

\begin{figure}
    \centering
    \includegraphics[width=\linewidth]{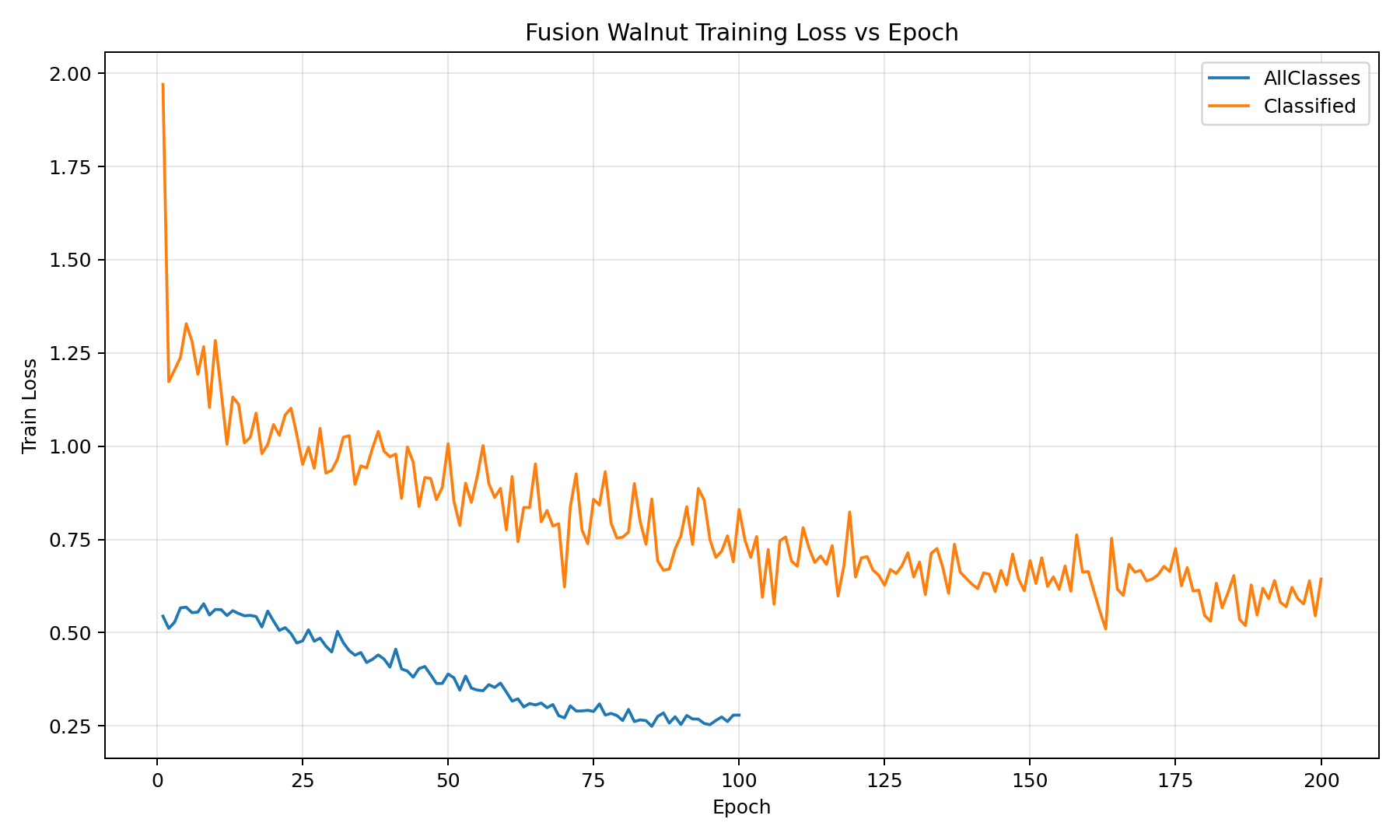}
    \caption{Learning Curve for the 2D-3D Fusion morphological model (Classified) and segmentation quality/object type model (AllClasses).}
\end{figure}

\end{document}

%% file: Pipeline_Fig.tex
\hspace*{-16pt}\begin{tikzpicture}[
  >=Latex,
  x=1.45cm, y=1.15cm,
  workflowstep/.style={inner sep=0pt, outer sep=0pt, font=\normalsize},
  data/.style={
    draw=datablue,
    fill=datablue!12,
    rounded corners,
    inner xsep=5pt,
    inner ysep=3.5pt,
    font=\normalsize
  }
]

% Central workflow steps
\node[workflowstep] (scan)       at (0.01,0)    {Scan};
\node[workflowstep] (extract)    at (2.0,0)  {Extract};
\node[workflowstep] (gui)        at (4.0,0)  {GUI};
\node[workflowstep] (classify)   at (6.0,0)  {Classify};
\node[workflowstep] (assemblage) at (8.0,0) {Assemblage};

% Data products
\node[data] (zstack)     at (0.75,1.25)   {z-stack};
\node[data] (pointcloud) at (2.5,1.25)   {Pointcloud};
\node[data] (rgb)        at (2.5,-1.25)  {RGB};
\node[data] (labels)     at (5.0,-1.25) {Labels};
\node[data] (prob)       at (7,-1.25)  {Probabilities};

% All edges behind nodes
\begin{pgfonlayer}{bg}

% Main workflow axis
\draw[->, thick] (scan) -- (extract);
\draw[->, thick] (extract) -- (gui);
\draw[->, thick] (gui) -- (classify);
\draw[->, thick] (classify) -- (assemblage);

% Scan -> z-stack -> Extract
\draw[->, thick] (scan.north) -- (zstack.west);
\draw[->, thick] (zstack.south east) -- (extract.north west);

% Extract -> RGB / Pointcloud
\draw[->, thick] (extract.north) -- (pointcloud.south);
\draw[->, thick] (extract.south) -- (rgb.north);

% RGB feeds
\draw[->, thick] (rgb.north east) -- (gui.south west);
\draw[->, thick] (rgb.east) -- (classify.south west);

% Pointcloud feeds
\draw[->, thick] (pointcloud.south east) -- (gui.north west);
\draw[->, thick] (pointcloud.east) -- (classify.north west);

% GUI -> Labels -> Classify
\draw[->, thick] (gui.south) -- (labels.north);
\draw[->, thick] (labels.north east) -- (classify.south);

% Classify -> Probabilities
\draw[->, thick] (classify.south) -- (prob.north);

% Probabilities -> GUI and Assemblage
\draw[->, thick] (prob.north west) -- (gui.south east);
\draw[->, thick] (prob.north east) -- (assemblage.south);

\end{pgfonlayer}

\end{tikzpicture}

%% file: Glossary_Table.tex
\begin{longtable}{p{0.24\linewidth}p{0.70\linewidth}}
\caption{Plain-language definitions of key technical terms used in the Sorometry workflow.}
\label{Glossary}\\
\toprule
\textbf{Term} & \textbf{Plain-language definition used in this paper} \\
\midrule
\endfirsthead

\toprule
\textbf{Term} & \textbf{Plain-language definition used in this paper} \\
\midrule
\endhead

\midrule
\multicolumn{2}{r}{\emph{Continued on next page}}\\
\midrule
\endfoot

\bottomrule
\endlastfoot

Artificial intelligence (AI)  Machine Learning (ML) & Computer methods that learn patterns from examples and then use those learned patterns to make predictions on new, unseen data. In this study, AI is used to help identify and summarize phytoliths automatically. \\

Sector & A \mbox{2 mm $\times$ 2 mm} sampled volume of a slide that is scanned and processed as one unit. Multiple sectors can be scanned from a single slide so that the slide can be sampled efficiently without always digitizing the entire slide. \\

z-stack (focus stack) & A series of images of the same sector taken at many different focal depths. Together, these images record how objects appear as the microscope focus moves through the settled material on the slide. \\

z-slice & One individual image plane within a z-stack, corresponding to a single focal depth. \\

Focus-stacked orthoimage (orthomosaic) & A single two-dimensional image made by combining the sharpest pixels from all z-slices, so that as much of the sector as possible appears in focus at once. \\

Point cloud & A three-dimensional representation of an object or surface made up of many points with x, y, and z coordinates, often also carrying RGB color values. In Sorometry, point clouds approximate the 3D form of segmented microscopic objects. \\

Segmentation & The automated step that separates a sector-level scan into individual object candidates. Here, segmentation produces one extracted object at a time, represented by a cropped image and a linked point cloud. \\

Segment & A single object candidate produced by segmentation. A segment may be a phytolith, a non-phytolith particle, debris, multiple objects stuck together, or a poorly extracted object. \\

Bounding box & The smallest rectangular crop that contains a segmented object in the image plane. It is used to cut out the corresponding 2D image for that object. \\

Labeling / annotation & The process in which a human expert assigns categories to a segment, such as its morphotype, whether it is well or poorly segmented, and whether it is a phytolith or another kind of particle. \\

Morphotype & A named morphological category based on shape and surface features. In this paper, morphotypes are the phytolith classes that the models are trained to recognize. \\

Diagnostic morphotype & A morphotype considered especially informative for identifying a particular plant or plant group. Such morphotypes may be rare, which is one reason why large-scale automated analysis is useful. \\

Training set & The subset of labeled examples used to teach a model which patterns correspond to which categories. \\

Train/test split & The division of labeled data into one subset used to fit the model (training) and another held back to evaluate how well the model performs on data it did not see during training. \\

Validation set & The subset of labeled examples that is hidden from the model until it's completely trained, to measure the empirical accuracy. \\

Convolutional neural network (CNN) & A class of machine-learning model designed to learn visual or structural patterns directly from data. In this study, CNN-based models are used to classify phytolith images, point clouds, or both together. \\

ConvNeXt & The image-based model family used here to classify phytoliths from their cropped 2D images. \\

PointNet & The point-cloud model family used here to classify phytoliths from their 3D point-cloud representations. Here, we use Pointnet++, but \textit{sorometry} supports the older Pointnet, and the transformer-based PointNeXt. \\

Fusion model & A model that combines information from both the 2D image and the 3D point cloud so that classification can use both surface appearance and three-dimensional shape. \\

Late-fusion scalar & An additional numerical input added near the end of the point-cloud model; here it records the log-transformed number of points in a segment so the model can take object scale/density into account. \\

Quality gate / pre-filter & A first-stage model that screens out debris, multicellular particles, and poorly segmented objects before the morphology model is applied. This reduces error by preventing the morphology classifier from being forced to classify unsuitable inputs. \\

Inference & The stage in which a trained model is run on unlabeled data to generate predictions. \\

Class probability & The probability a model assigns to each possible class for a given segment. The highest of these probabilities is the model's top prediction. \\

Maximum-probability class & The class receiving the highest predicted probability for a segment; this is the category usually reported as the model's predicted label. \\

Confidence & The strength of the model's support for a prediction, usually expressed through the class probabilities. Higher confidence means the model more strongly favors one class over the alternatives. \\

Compositional data & Data that describe parts of a whole, such as the relative abundances of predicted morphotypes within a sector or slide. These data require special statistical treatment because the parts are not independent and must sum to a total. \\

Principal components analysis (PCA) & A dimensionality-reduction method that summarizes the main axes of variation among samples so they can be visualized in a lower-dimensional space. \\

Hierarchical clustering & A method that groups samples according to similarity, producing a dendrogram that shows which samples are more similar to each other in composition. \\

Chi-square test of independence & A statistical test used here to ask whether the distribution of predicted morphotypes differs more than expected by chance across sectors from the same slide. \\

Bayesian finite mixture of multinomials & A statistical model that estimates what combination of reference plant sources could plausibly have generated the observed counts of morphotypes in an archaeological sample. \\

Posterior distribution & In Bayesian analysis, the range of parameter values that remain plausible after combining prior assumptions with the observed data. Here, it expresses uncertainty about the contribution of different source plants to a sample. \\

Reference collection & A set of slides from known plant material used as a baseline for training, comparison, and mixture modeling. \\

\end{longtable}

%% file: Confusion_Supplement.tex
\hspace*{-128pt}

\begin{figure}[b!]
\centering
\includegraphics[width=\linewidth]{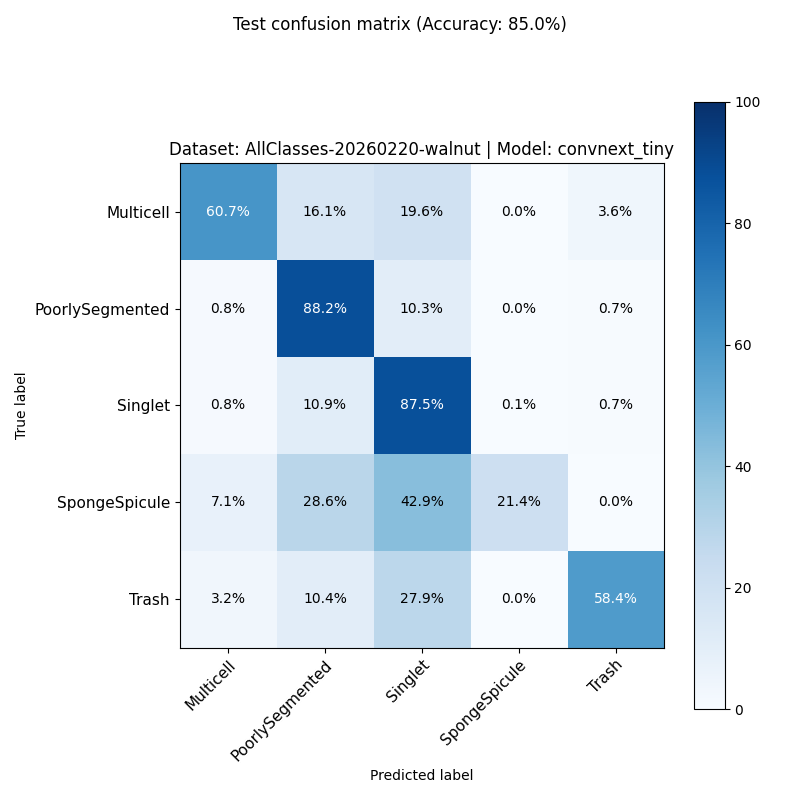}
\caption{Segmentation quality and object type classification performance for the ConvNeXt 2D image model, showing percent true values classified as a given predicted value.}
\label{AllClasses_Image}
\end{figure}

\begin{figure}
\centering
\includegraphics[width=\linewidth]{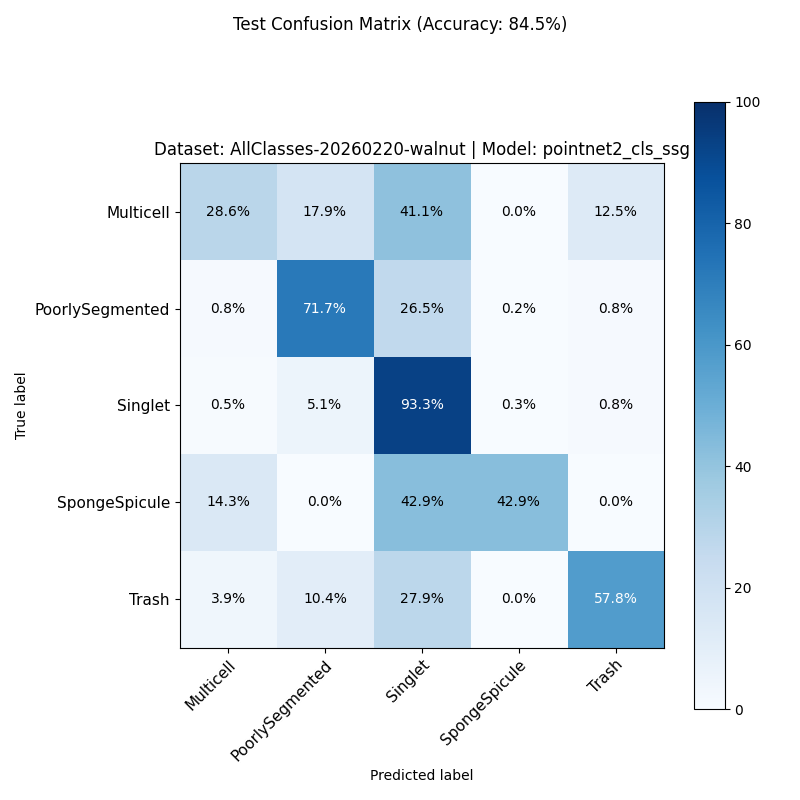}
\caption{Segmentation quality and object type classification performance for the PointNet++ 3D pointcloud model with late  fusion scalar, showing percent true values classified as a given predicted value.}
\label{AllClasses_Pointcloud}
\end{figure}

\begin{figure}
\centering
\includegraphics[width=\linewidth]{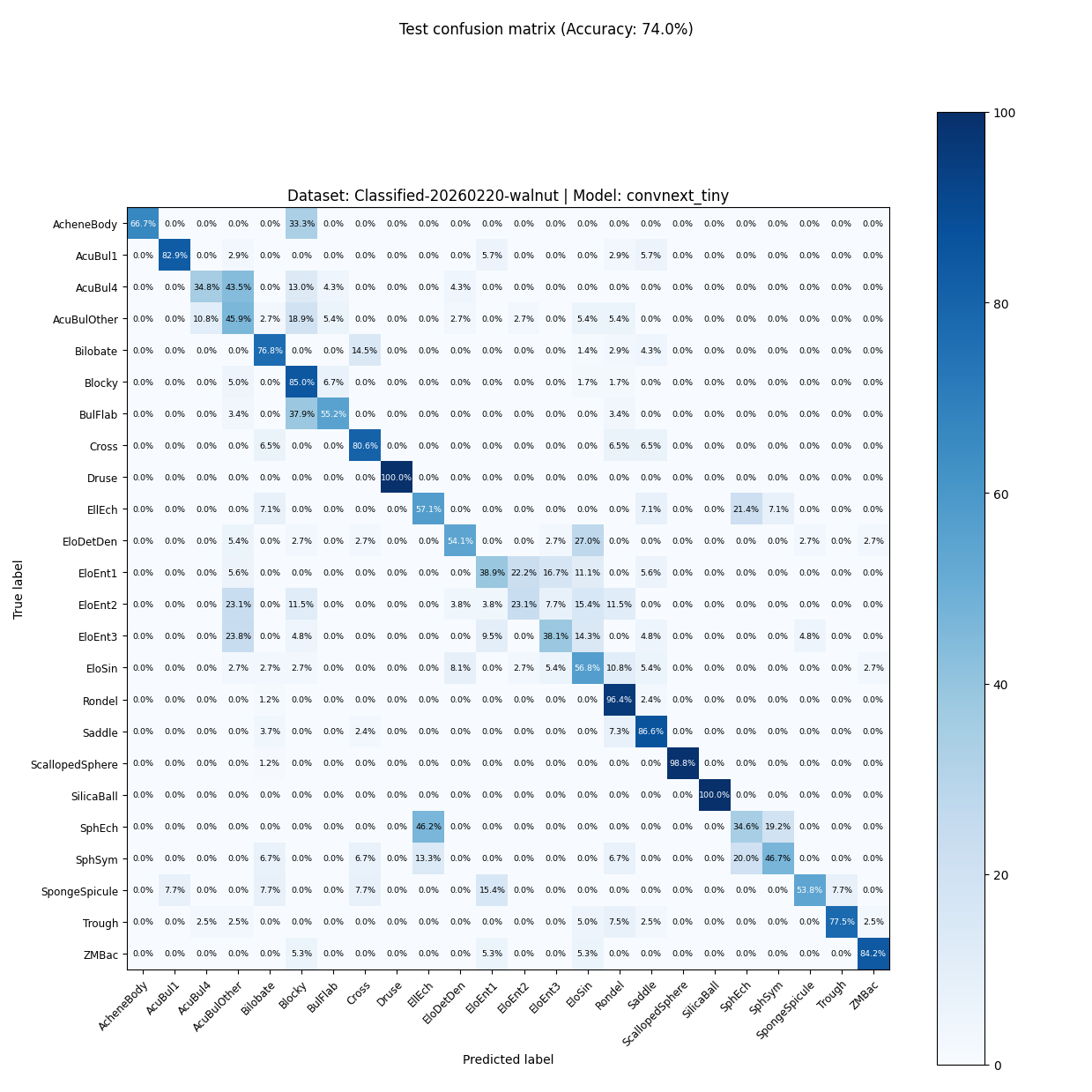}
\caption{Morphotype classification performance for the ConvNeXt 2D image model, showing percent true values classified as a given predicted value.}
\label{Classified_Image}
\end{figure}

\begin{figure}
\centering
\includegraphics[width=\linewidth]{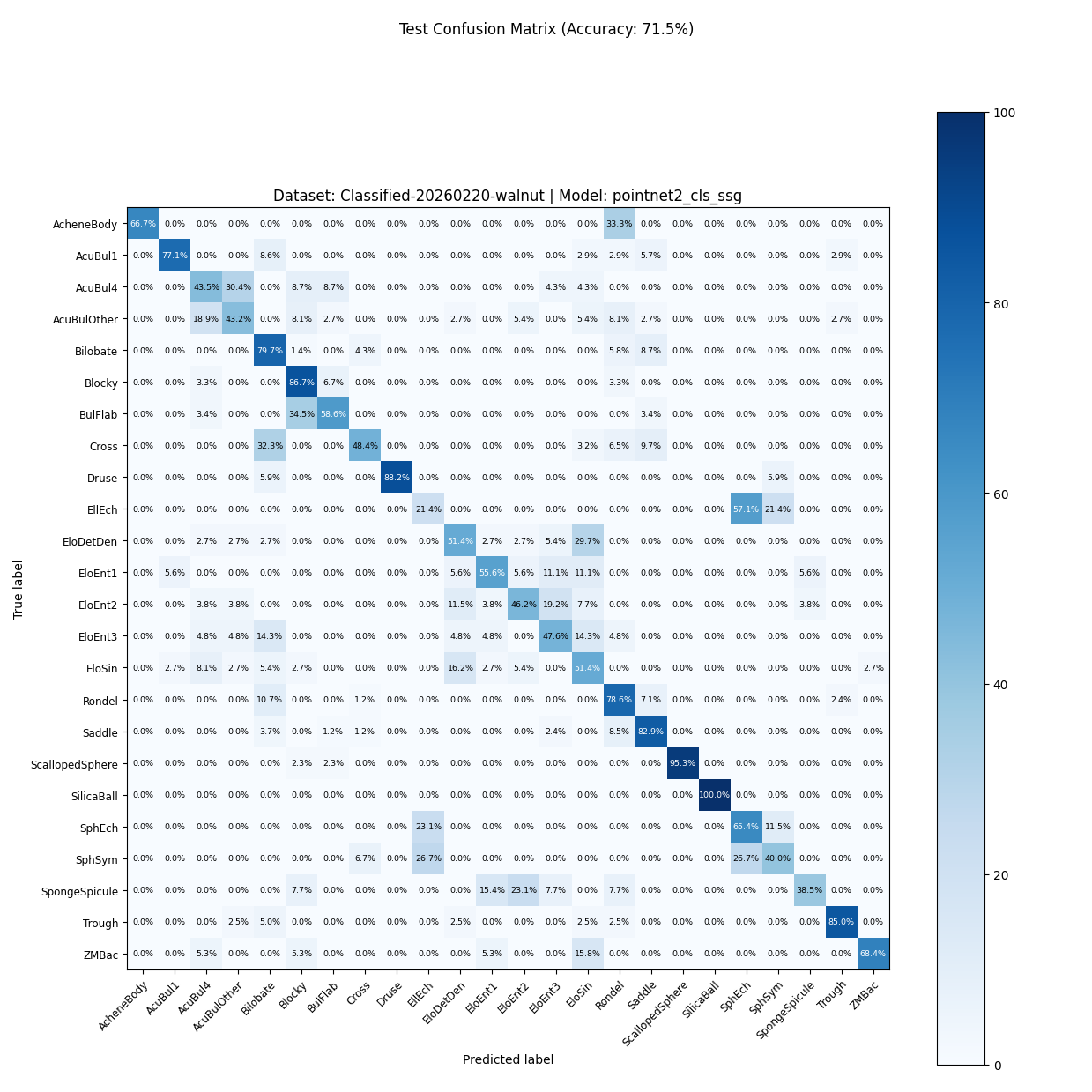}
\caption{Morphotype classification performance for the Pointnet++ 3D pointcloud model with late fusion scalar, showing percent true values classified as a given predicted value.}
\label{Classified_Pointcloud}
\end{figure}